%% file: main.tex
\documentclass[10pt,letterpaper,twocolumn]{article}
\usepackage[pagenumbers]{cvpr} % To force page numbers, e.g. for an arXiv version

\definecolor{cvprblue}{rgb}{0.21,0.49,0.74}
\usepackage[pagebackref,breaklinks,colorlinks,allcolors=cvprblue]{hyperref}
\usepackage{bm}
\usepackage{amsmath}
\usepackage{graphicx}
\usepackage[linesnumbered, ruled, lined]{algorithm2e}
\usepackage{multirow}
\usepackage{multicol}
\usepackage{placeins}

\usepackage{cuted}      
\usepackage{caption} 
\usepackage{float}

\usepackage{algorithmic}
\usepackage{amssymb}
\newcommand{\COMMENT}[1]{\hfill \(\triangleright\) #1} % 改进注释格式

\DeclareFixedFont{\ttb}{T1}{txtt}{bx}{n}{8} % for bold
\DeclareFixedFont{\ttm}{T1}{txtt}{m}{n}{8}  % for normal

\usepackage{color}
\definecolor{deepblue}{rgb}{0,0,0.5}
\definecolor{deepred}{rgb}{0.6,0,0}
\definecolor{deepgreen}{rgb}{0,0.5,0}

\def\paperID{6148} % *** Enter the Paper ID here
\def\confName{CVPR}
\def\confYear{2025}

\newcommand{\mymethod}{EmbodiedGen}

\title{{\mymethod}: Towards a Generative 3D World Engine \\for Embodied Intelligence}
\author{
Xinjie Wang$^{1}$ \quad
Liu Liu$^{1}$ \quad
Yu Cao$^{2}$ \quad
Ruiqi Wu$^{5, 1}$ \quad
Wenkang Qin$^{2}$ \\
Dehui Wang$^{4, 3}$ \quad
Wei Sui$^{3}$ \quad 
Zhizhong Su$^{1}$ \\
$^1$Horizon Robotics \quad $^2$GigaAI \quad $^3$D-Robotics \\
$^4$Shanghai Jiao Tong University \quad $^5$VCIP, CS, Nankai University\\
}
\begin{document}

\twocolumn[{
\renewcommand\twocolumn[1][]{#1}
\maketitle
\begin{center}
    \vspace*{-2mm}
    \captionsetup{type=figure}
    \includegraphics[width=\textwidth]{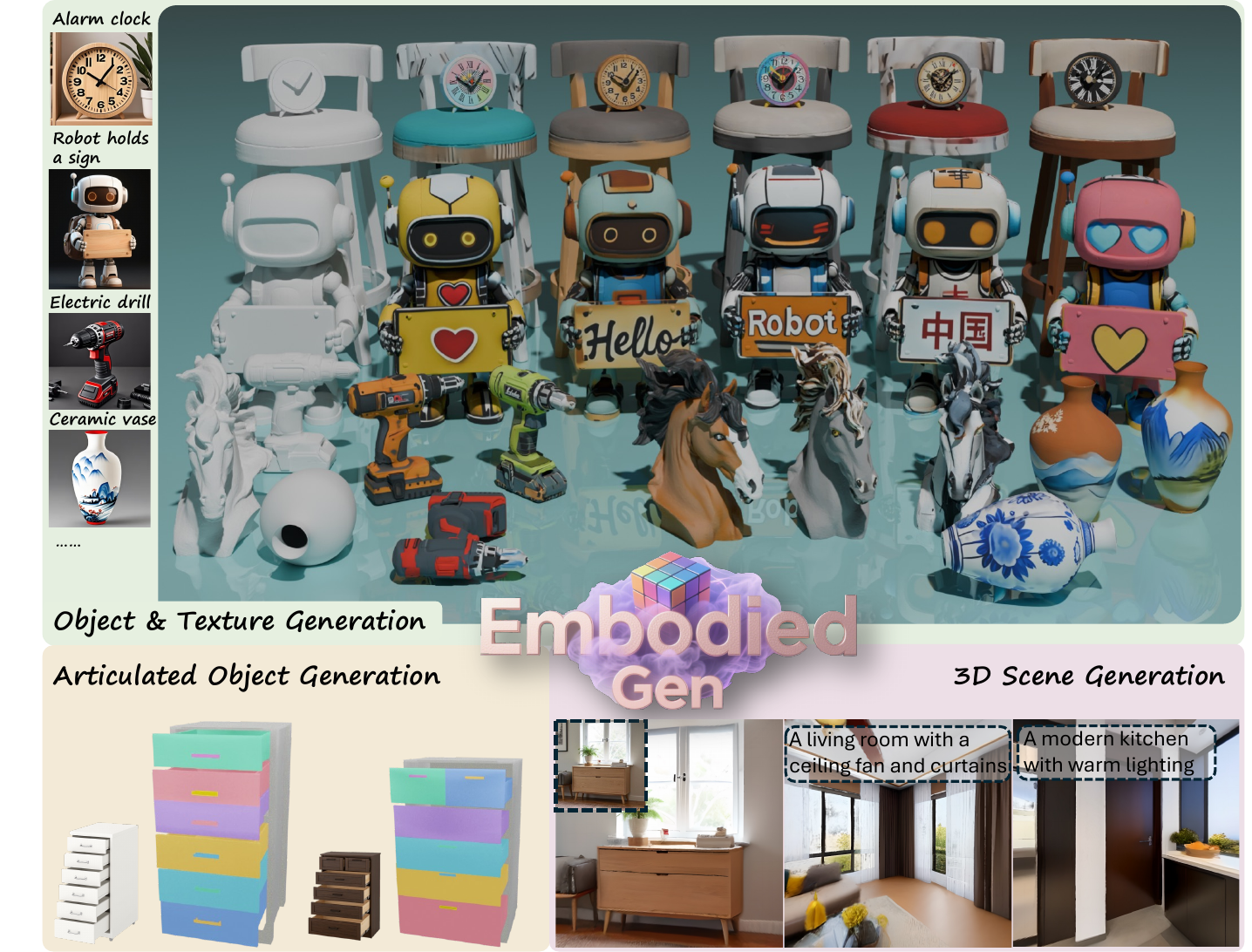}
    \captionof{figure}{\textbf{\mymethod}, a toolkit for embodied intelligence interactive 3D world generation. {\mymethod} enables controllable generation of rigid and articulated assets with accurate real-world scale and physical properties, along with stylistically diverse background generation and visually rich texture generation and editing. These assets can be seamlessly integrated into various simulators such as OpenAI Gym\cite{openai_gym}, Isaac Lab\cite{mittal2023orbit}, MuJoCo\cite{todorov2012mujoco} and SAPIEN\cite{Xiang_2020_SAPIEN}. These capabilities form a foundation for digital twinning, large-scale data augmentation and embodied intelligence tasks such as manipulation and navigation across a wide range of simulation environments.}
    \label{fig:results}
    \vspace*{-10mm}
\end{center}
}]

% \columnbreak
\clearpage

\input{sec/0_abstract}

\input{sec/1_intro}
\input{sec/2_main}

{
    \small
    \bibliographystyle{ieeenat_fullname}
    \bibliography{main}
}

% \setcounter{section}{0}  % 重置section编号

\end{document}

%% file: sec/0_abstract.tex
\begin{abstract}
Constructing a physically realistic and accurately scaled simulated 3D world is crucial for the training and evaluation of embodied intelligence tasks. 
The diversity, realism, low cost accessibility and affordability of 3D data assets are critical for achieving generalization and scalability in embodied AI.
However, most current embodied intelligence tasks still rely heavily on traditional 3D computer graphics assets manually created and annotated, which suffer from high production costs and limited realism. These limitations significantly hinder the scalability of data driven approaches.
We present \textbf{\textit{\mymethod}}, a foundational platform for interactive 3D world generation. 
It enables the scalable generation of high-quality, controllable and photorealistic 3D assets with accurate physical properties and real-world scale in the Unified Robotics Description Format (URDF) at low cost.
These assets can be directly imported into various physics simulation engines for fine-grained physical control, supporting downstream tasks in training and evaluation. {\mymethod} is an easy-to-use, full-featured toolkit composed of six key modules:
\textit{Image-to-3D}, \textit{Text-to-3D}, \textit{Texture Generation}, \textit{Articulated Object Generation}, \textit{Scene Generation} and  \textit{Layout Generation}. \textbf{\textit{\mymethod}} generates diverse and interactive 3D worlds composed of generative 3D assets, leveraging generative AI to address the challenges of generalization and evaluation to the needs of embodied intelligence related research.
\textbf{\textit{\mymethod}}\footnote{\url{https://horizonrobotics.github.io/robot_lab/embodied_gen/index.html}} are publicly available to foster future research.

\end{abstract}

% \textbf{Image-to-3D}: generates physically realistic 3D object assets from a single image, facilitating the creation of digital twins.
% \textbf{Text-to-3D}: generates 3D object assets from text descriptions, supporting low-cost high-quality data augmentation.
% \textbf{Texture Generation}: facilitates texture generation and editing for 3D object assets, offering highly controllable and multi-style visual feature editing.
% \textbf{Articulated Object Generation}: generates manipulative articulated 3D objects from a dual-state image pair or the text description.
% \textbf{Scene Generation}: generates diverse 3D scene backgrounds, supporting the generation of varied environmental contexts.
% \textbf{Layout Generation}: generates a diverse and interactive 3D world composed of generative 3D assets, driven by the above modules and guided by the given task description.

%% file: sec/1_intro.tex
\section{Introduction}
\label{sec:intro}

Despite the remarkable success of foundation models such as CLIP~\cite{clip2021} and GPT~\cite{radford2019language, achiam2023gpt}, which leverage large-scale internet data, extending this paradigm to the needs of embodied intelligence related research presents significant challenges. Data collection for embodied AI tasks is substantially more expensive and constrained, often requiring real world interaction, involving complex physical dynamics, making each data point several orders of magnitude more costly. Moreover, robotic data is often context-specific and non-transferable across tasks or embodiments, severely limiting reusability and scalability. Achieving general-purpose embodied intelligence in the physical world requires techniques such as digital twins, simulation-based augmentation, and reinforcement learning in physically realistic environments. These goals demand access to large-scale, diverse, and high-quality 3D asset libraries, as well as efficient pipelines for rapidly constructing interactive 3D environments.
Recent advances in diffusion based generative models~\cite{ho2020denoising, nichol2021improved, ldm, wu2024lamp} and 3D asset generation~\cite{hong2024lrm, charatan2024pixelsplat, liu2023zero1to3zeroshotimage3d, poole2022dreamfusiontextto3dusing2d, zhang2024claycontrollablelargescalegenerative, zhao2025hunyuan3d20scalingdiffusion} have sparked growing interest in bridging this gap. However, existing 3D generation toolkit often fall short for robotics applications, as conventional graphics assets typically lack physical realism, watertight geometry and accurate scale, leading to unreliable collision modeling and unrealistic interactions in simulators.
We introduce \textbf{\textit{{\mymethod}}}, a toolkit for interactive 3D world generation, enables low-cost, high-quality, and highly controllable asset generation in URDF, complete with watertight geometry and physically plausible properties. Our main contributions are:

\begin{enumerate}
% \begin{itemize}[leftmargin=*]
\item \textbf{Toolkit for interactive 3D world generation.} \textit{\mymethod} is the first comprehensive toolkit for generating interactive 3D worlds to the needs of embodied intelligence related research. It supports real-to-sim digital twin creation and enables the controllable generation of diverse 3D assets, which can be seamlessly imported into simulators.
\item \textbf{Physically accurate, simulator-ready assets with high fidelity.}
\textit{\mymethod} generates assets that not only achieve state-of-the-art visual quality but are also physically plausible and ready for direct use in simulation environments. Each asset is enriched with physical properties, inspection metadata, textual descriptions, watertight geometry and dual representations in both 3D Gaussian Splatting (3DGS) and mesh formats.
\item \textbf{Accessible and open-source.} We release easy-to-use, open-sourced pipelines and services to facilitate community development and research in embodied intelligence.
% \end{itemize}
\end{enumerate}

%% file: sec/2_main.tex
\begin{figure*}[!ht]
\centering
\includegraphics[width=\linewidth]{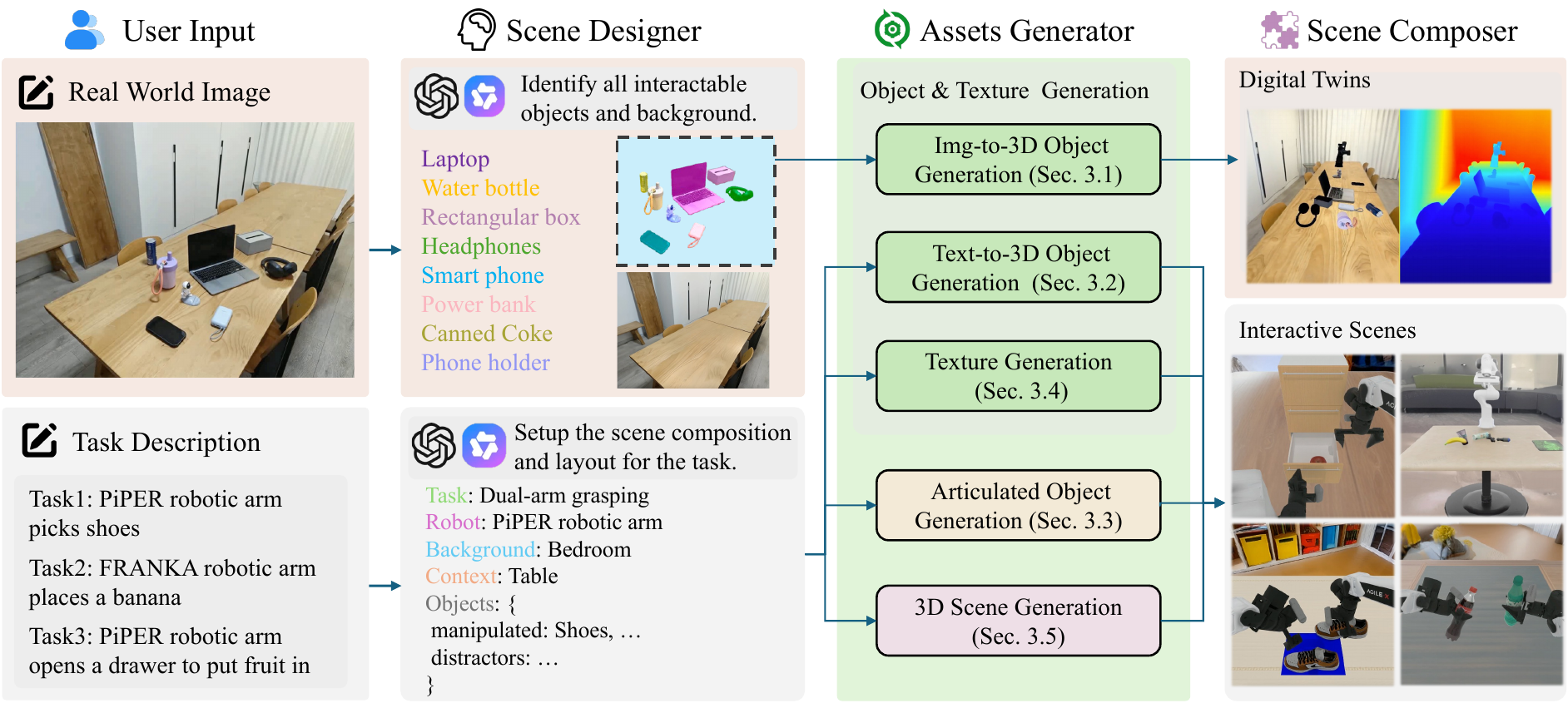}
\vspace*{-8mm}
\caption{The framework of \textit{{\mymethod}}. 
It enables the creation of a digital twin within a simulation environment from a single image. Alternatively, given a task description, \textit{{\mymethod}} autonomously generates the scene layout, synthesizes detailed 3D object assets, and arranges them in semantically and physically plausible configurations. This facilitates the effortless construction of an interactive 3D world, supporting a wide range of embodied intelligence related research in diverse virtual environments.}
\vspace*{-3mm}
\label{fig:gen_engine}
\end{figure*}

\section{Related Work}
\label{sec:rel_work}

\paragraph{3D Asset Generation}
The goal of 3D object generation is to produce a corresponding 3D representation from an input image or a textual description. Existing approaches to this task can be broadly categorized into three representative paradigms: feedforward generation, optimization-based generation, and view reconstruction.
Feedforward generation leverages large models to produce a 3D representation of the input prompt in a single forward pass. This category includes methods such as LRM\cite{hong2024lrm}, PixelSplat\cite{charatan2024pixelsplat}, GRM\cite{xu2024grm}, and MVSplat\cite{Chen_2024_mvsplat}, which are notable for their inference time efficiency.
Optimization-based generation, as exemplified by methods like DreamFusion\cite{poole2022dreamfusion} and DreamMat\cite{zhang2024dreammat}, directly optimizes the parameters of the 3D representation using score distillation sampling (SDS) guided by diffusion models and differentiable rendering. This often results in higher-quality outputs at the cost of increased computation time.
View reconstruction methods generate multi-view 2D images and reconstruct the final 3D representation via sparse-view geometry. Representative works in this line include Zero123\cite{liu2023zero1to3}, Unique3d\cite{wu2024unique3d}, MVDream\cite{shi2024mvdream}, and MV-Adapter\cite{huang2024mvadapter}.
Driven by the demand for higher-quality 3D objects, recent methods such as CLAY\cite{zhang2024clay}, Hunyuan3D\cite{hunyuan3d22025tencent}, Meta3DGen\cite{bensadoun2024meta3dgen}, and Trellis\cite{xiang2024structured} have adopted a decoupled pipeline separating geometry and texture generation into two stages, followed by texture reprojection to fuse geometry with realistic textures.
Beyond rigid object generation, methods such as URDFormer\cite{chen2024urdformerpipelineconstructingarticulated} and SINGAPO\cite{liu2025singaposingleimagecontrolled} have been proposed to generate articulated objects.
However, these methods are primarily limited to graphics-centric object generation. The resulting objects lack real-world scale and physical properties, and there is no guarantee of watertightness or geometric integrity. These limitations significantly hinder their direct applicability in physics-based simulators. 

\paragraph{3D Scene Generation}
Recent methods like LucidDreamer\cite{chung2023luciddreamerdomainfreegeneration3d} adopt 3DGS\cite{kerbl20233dgaussiansplattingrealtime} for flexible and consistent scene rendering, but are mainly limited to forward facing views. To enable full 360° scene generation, panoramic representations have been explored. PERF\cite{wang2023perfpanoramicneuralradiance} pioneered panoramic NeRFs for novel view synthesis from a single panorama. DreamScene360\cite{zhou2024dreamscene360unconstrainedtextto3dscene}, HoloDreamer\cite{zhou2024holodreamerholistic3dpanoramic} and WorldGen\cite{worldgen2025ziyangxie} extended this with panoramic Gaussian splatting, while LayerPano3D\cite{yang2025layerpano3dlayered3dpanorama} introduced layered panoramas that are lifted into 3D splatting for handling complex scenes. However, these methods are limited to generating static 3D scenes without interactivity, making them unsuitable for the requirements of embodied intelligence related research.

\paragraph{Embodied Intelligence Tasks}
Prior works such as RoboTwin\cite{mu2025robotwin}, Gen2Sim\cite{katara2023gen2sim}, MatchMaker\cite{wang2025matchmaker} and ACDC\cite{dai2024automatedcreationdigitalcousins} have explored using 3D generation techniques to augment asset libraries within simulators. However, due to limitations in the quality and efficiency of 3D generation, the diversity of assets remains limited, and the environments are often restricted to simplistic backgrounds, which are insufficient for large-scale data generation and evaluation in embodied intelligence tasks. To address these challenges, we propose \textit{{\mymethod}}, a data-centric foundation for embodied AI, enables the generation of diverse object and background assets from either images or text prompts, and supports texture editing for enhanced visual richness. This framework effectively supports real-to-sim transfer, data augmentation, and physics-based simulation in different simulators\cite{todorov2012mujoco,openai_gym,mittal2023orbit,Xiang_2020_SAPIEN}, accelerating the development of embodied intelligence systems.

\begin{figure*}[!htbp]
\centering
\includegraphics[width=\linewidth]{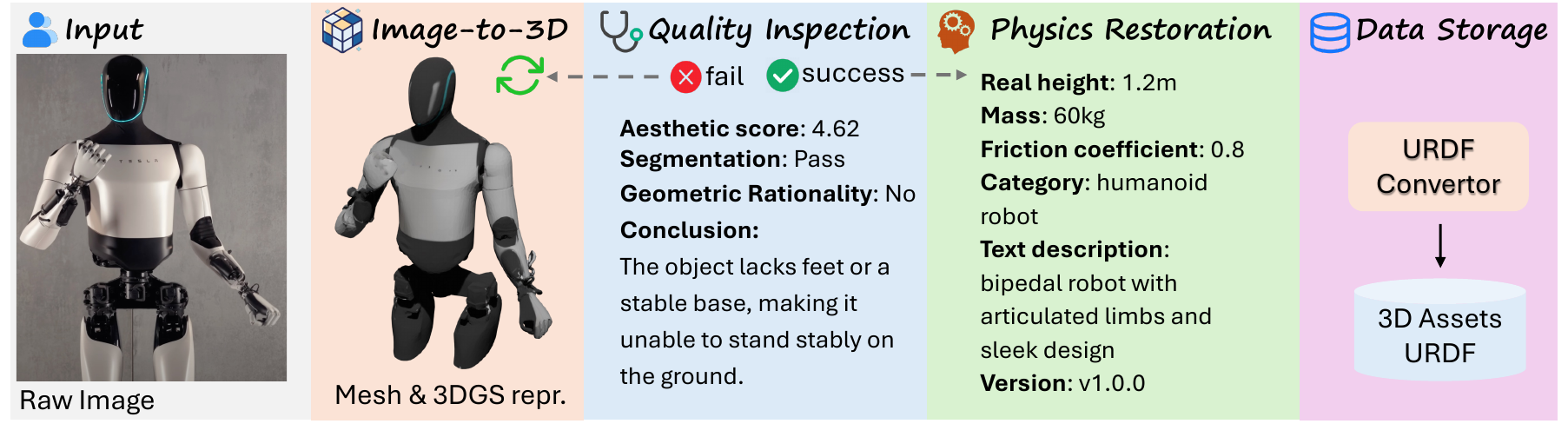}
\vspace*{-8mm}
\caption{
Overview of \textit{\mymethod} Image-to-3D Pipeline. From a single image, the system generates mesh and 3DGS assets, conducts automatic quality inspectioin (aesthetics, segmentation, geometry), and re-generate failed outputs by auto-adjusted settings. A physics expert module restores real-world scale and physical semantics, and the assets are saved in URDF format.
}
\vspace*{-3mm}
\label{fig:image_pipe}
\end{figure*}

\section{Generative 3D World Engine}
\label{sec:gene_engine}

We present \textbf{\textit{\mymethod}}, a novel framework for generating interactive 3D worlds to the needs of embodied intelligence related research. Leveraging generative AI, our approach enables the creation of diverse, customizable environments that support the development and evaluation of embodied agents (see Figure~\ref{fig:gen_engine}).

% \textbf{Image-to-3D}: generates physically realistic 3D object assets from a single image, facilitating the creation of digital twins.
% \textbf{Text-to-3D}: generates 3D object assets from text descriptions, supporting low-cost high-quality data augmentation.
% \textbf{Texture Generation}: facilitates texture generation and editing for 3D object assets, offering highly controllable and multi-style visual feature editing.
% \textbf{Articulated Object Generation}: generates manipulative articulated 3D objects from a dual-state image pair or the text description.
% \textbf{Scene Generation}: generates diverse 3D scene backgrounds, supporting the generation of varied environmental contexts.
% \textbf{Layout Generation}: generates a diverse and interactive 3D world composed of generative 3D assets, driven by the above modules and guided by the given task description.

% In the remainder of this section, we first describe the 3D object generation modules given image and text inputs in Sec.\ref{sec:robo_asset_gen_img} and Sec.\ref{sec:robo_asset_gen_text}, respectively. We then introduce the articulated object generation in Sec.\ref{sec:articulated_gen}, the texture generation and editing module in Sec.\ref{sec:robo_asset_gen_edit}, and the 3D scene generation in Sec.~\ref{sec:robo_scene_gen}.

In this section, we first present the 3D object generation modules, including Image-to-3D, which generates physically realistic 3D object assets from a single image to facilitate digital twin creation (Sec.\ref{sec:robo_asset_gen_img}), and Text-to-3D, which generates 3D objects from text descriptions for low cost, high-quality data augmentation (Sec.\ref{sec:robo_asset_gen_text}). We then introduce Articulated Object Generation (Sec.\ref{sec:articulated_gen}), which produces manipulable articulated 3D assets from either dual-state image pairs or text descriptions. Texture Generation is described in Sec.\ref{sec:robo_asset_gen_edit}, enabling highly controllable and multi-style texture editing for 3D assets. Finally, we present Scene Generation (Sec.~\ref{sec:robo_scene_gen}), which generates diverse background environments and supports the composition of interactive 3D scenes.

\subsection{Image-to-3D Object Generation}
\label{sec:robo_asset_gen_img}

\paragraph{Method Overview}
The capabilities of community-driven 3D object asset generation are rapidly advancing and are expected to continue improving. To fully leverage this progress, we focus on building an image-to-3D system to the needs of embodied intelligence related research. For the model component, we leverage open-source models. This approach ensures that our image-to-3D capabilities can be easily extended as community models improve. Specifically, we use Trellis\cite{xiang2024structured} due to its superior geometric generation quality and its ability to produce consistent 3D representations in both mesh and 3DGS\cite{kerbl20233dgaussiansplattingrealtime} formats. However, Trellis has several limitations that hinder its direct use in embodied AI tasks: the generated textures exhibit poor visual quality, particularly due to excessive highlights that result in noticeable whitening when baked onto the mesh. Additionally, the resulting files are purely graphical assets without real-world scale, physical properties, or physically plausible geometry, making them unsuitable for direct use in physics simulators\cite{mittal2023orbit,Xiang_2020_SAPIEN,todorov2012mujoco}. We focus on three key improvements: (1) developing a complete data twinning pipeline for embodied intelligence asset generation, capable of producing data assets with realistic properties, accurate scale, and physically consistent watertight geometry that can be directly imported into simulation engines; (2) enhancing texture quality by applying highlight removal and super-resolution, resulting in high-quality, high-resolution textures; (3) developing a diffusion-based model for articulated 3D object generation to meet the growing demand for complex data assets in diverse simulation tasks.

\paragraph{Physically Realistic 3D Asset Generation}
As illustrated in Figure~\ref{fig:image_pipe}, we leverage Trellis\cite{xiang2024structured} to generate 3D representations of input images. We further employ GPT-4o\cite{achiam2023gpt} and Qwen\cite{qwen} to build a physics expert agent. Specifically, the agent estimates the real-world height of the asset by rendering a frontal view of the generated object and applying text prompt constraints. Given that width, length, and height are interdependent, scaling the height enables accurate recovery of the mesh and 3DGS's true dimensions. For assets with inherent ambiguity in size, a text-guided physical property restoration interface is available, allowing users to specify context (e.g., “a tiger plush toy” or “a tiger animal”) for more accurate size prediction. Given four orthographic views of a rendered 3D asset as input, the physics expert agent can further estimate physical properties such as friction coefficient and mass, associate them with semantic descriptions, and categorize the object 3D assets accordingly.

\begin{figure}[!htbp]
\centering
\includegraphics[width=\linewidth]{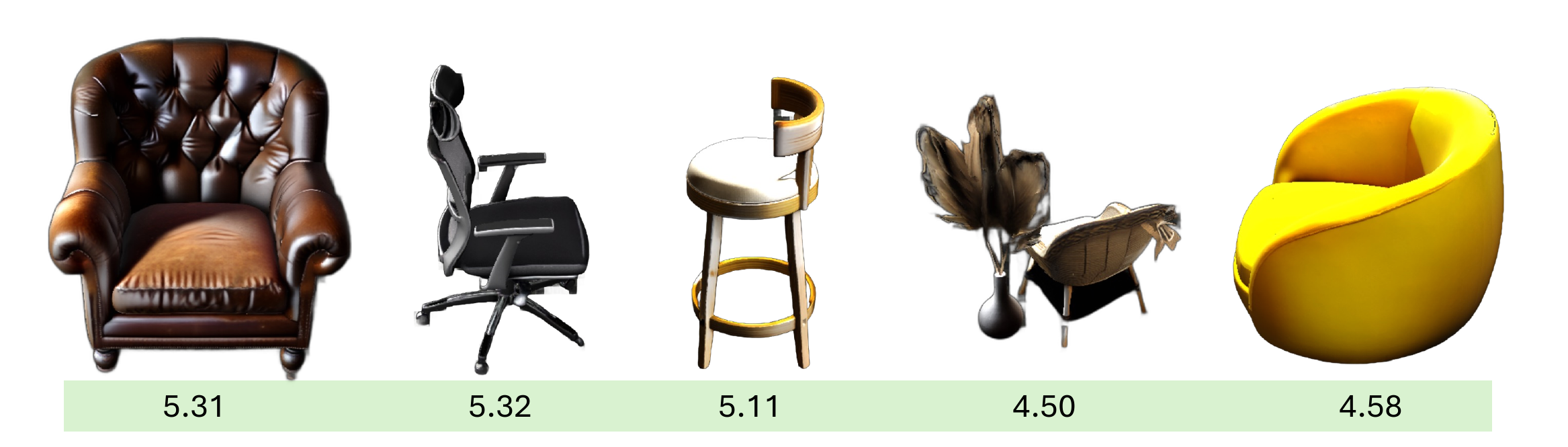}
\vspace*{-7mm}
\caption{\textit{AestheticChecker} is used to evaluate the texture quality of generated assets. Assets displaying richer texture details receiving higher scores.}
\vspace*{-3mm}
\label{fig:a_score}
\end{figure}

\begin{figure}[!htbp]
\centering
\includegraphics[width=\linewidth]{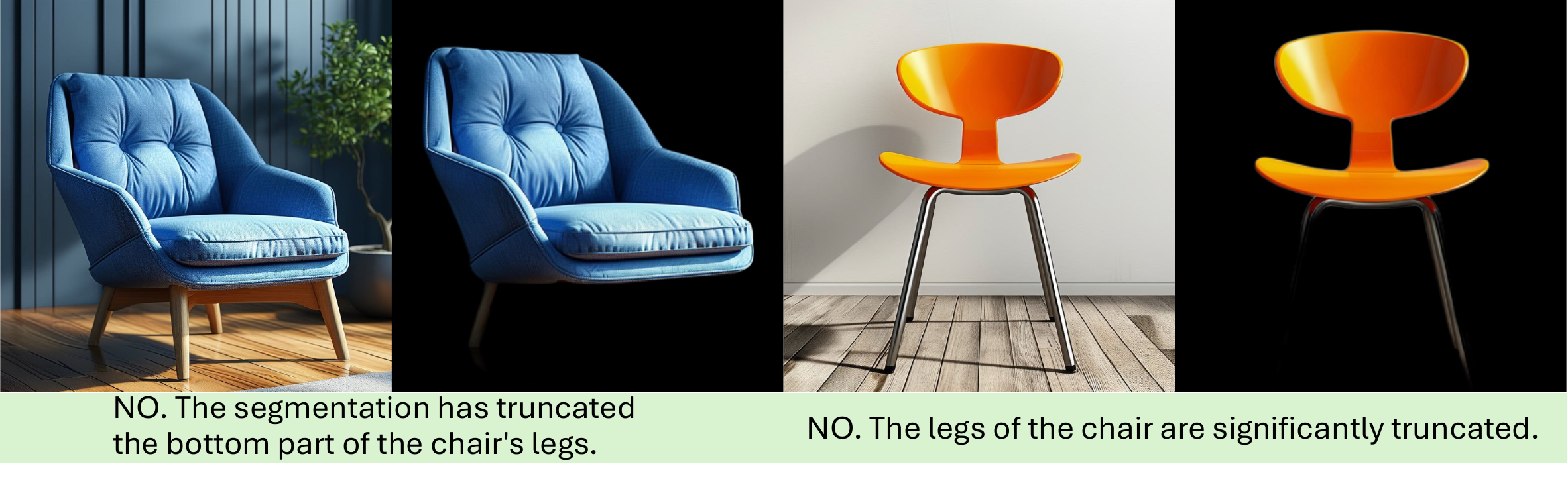}
\vspace*{-7mm}
\caption{Examples of segmentation failure cases automatically filtered by \textit{ImageSegChecker}.}
\vspace*{-3mm}
\label{fig:seg_check}
\end{figure}

\begin{figure}[!htbp]
\centering
\includegraphics[width=\linewidth]{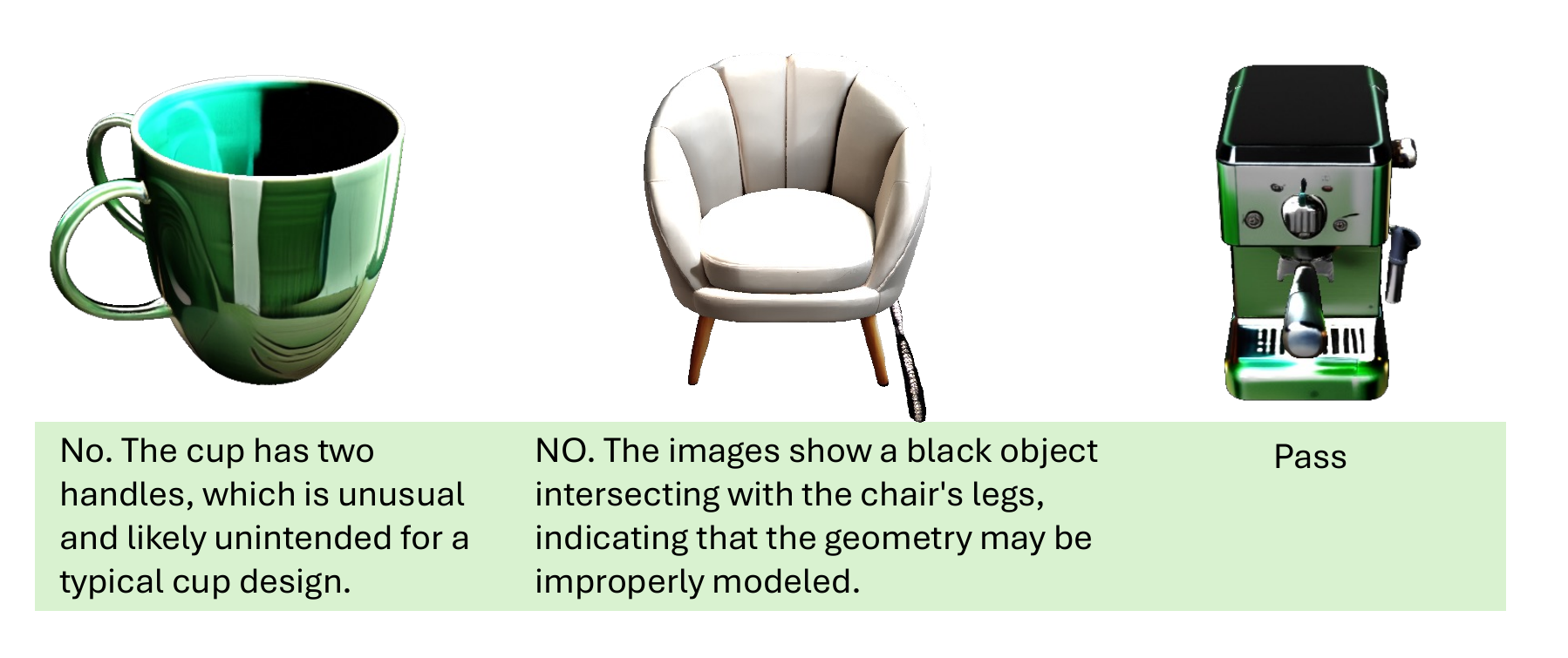}
\vspace*{-7mm}
\caption{Examples of geometric rationality inspection by \textit{MeshGeoChecker}.}
\vspace*{-3mm}
\label{fig:geo_check}
\end{figure}

\paragraph{Automated Quality Inspection} We develop an automated quality inspection module, utilizing the \textit{AestheticChecker}~\cite{schuhmann_aesthetic_2025} as a measure of visual quality, as it has a positive correlation with texture richness, see Figure~\ref{fig:a_score}. We found that the quality of foreground segmentation has a significant impact on the quality of 3D asset generation, so we further build a \textit{ImageSegChecker} using GPT-4o for foreground extraction quality assessment, see Figure~\ref{fig:seg_check}. To ensure robust segmentation quality across different domains, we provide three different foreground segmentation models, SAM\cite{kirillov2023segany}, REMBG\cite{Gatis_rembg_2025}, RMBG14\cite{briaai_rmbg_1_4}. If \textit{ImageSegChecker} detects a segmentation failure, the system switches to an alternative model for retry. A \textit{MeshGeoChecker} inspects the asset by rendering four orthogonal views and assessing geometric completeness and rationality,  see Figure~\ref{fig:geo_check}. Assets that pass the quality inspection are converted into URDF format and stored. Those that fail any stage of the pipeline are sent back to the corresponding generation step using adjusted settings and seeds.

\begin{figure}[!htbp]
\centering
\includegraphics[width=\linewidth]{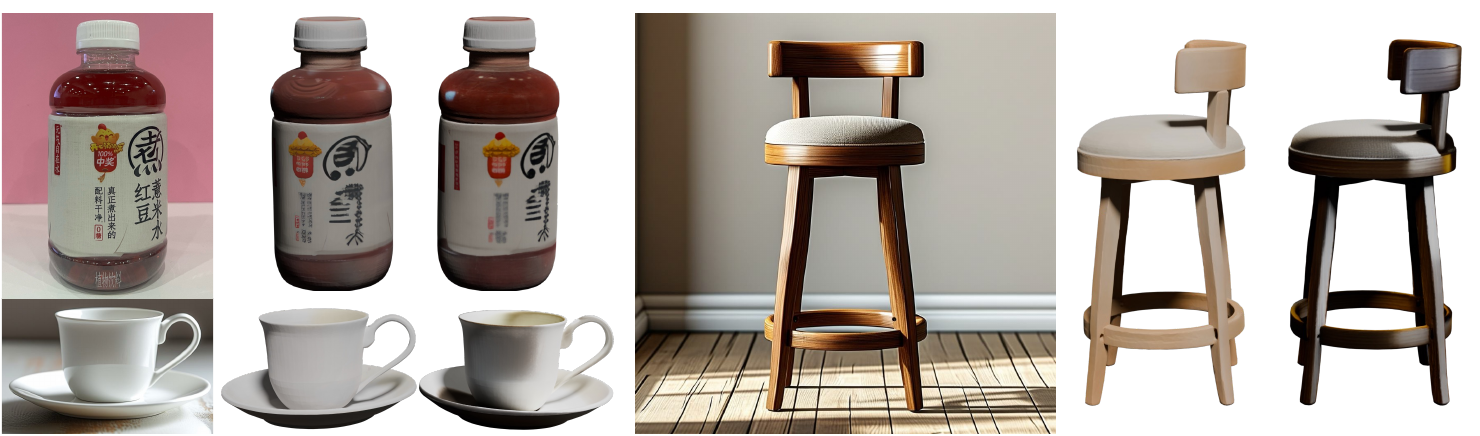}
\vspace*{-7mm}
\caption{From left to right: the original image, the result of our optimized texture back-projection, and the result using Trellis’s original texture back-projection. Our method effectively mitigates the influence of highlights and shadows on the mesh texture while producing significantly sharper texture details.}
\vspace*{-3mm}
\label{fig:texture}
\end{figure}

\paragraph{Texture Back Projection Optimization}
Unlike methods like Trellis that rely on optimization-based baking of multi-view RGB images rendered from 3DGS\cite{kerbl20233dgaussiansplattingrealtime} back into 3D space, our optimized approach employs a geometry-determined projection scheme fused with view normals. Before re-projecting textures back into 3D space, we apply global highlight removal and super-resolution to the RGB images, resulting in high-quality, 2K resolution texture UV maps. Specifically, we use a delighting model\cite{hunyuan3d22025tencent} to remove lighting effects from the multi-view textures while maintaining consistent style and brightness across views. We further apply Real-ESRGAN~\cite{wang2021realesrgan} to independently perform 4x super-resolution on each view, enhancing the resolution to 2048x2048. Our experiments show that independent super-resolution for each view does not compromise the consistency or quality of the final 3D asset texture. The algorithmic workflow is illustrated in Algorithm~\ref{alg:back_project}. 
We present in Figure~\ref{fig:texture} the improvement in mesh texture quality achieved by the optimized texture back-projection module.

\begin{algorithm}
\caption{Compute Texture by Multi-View Color Images Back-Projection}
\label{alg:back_project}
\KwIn{
    $I\in\mathbb{R}^{N\times H_0\times W_0\times 3}$: multi-view color images;\\
    $\mathcal{M} = (V, F)$: input mesh, where\\
    \quad $V$ is mesh vertices, $F$ is mesh faces;\\
    $W\in \mathbb{R}^{N}$: view confidence weights;\\
    $\theta$: angle threshold (default: $70^\circ$).
}
\KwOut{
  $T\in\mathbb{R}^{H_{\mathrm{tex}}\times W_{\mathrm{tex}}\times 3}$: texture uv map.\\
}
\BlankLine

\SetKwBlock{Delighting}{1. Delighting Color Image}{}
\Delighting{
    $I_{\text{grid}} \in \mathbb{R}^{(N \times H_0) \times W_0 \times 3} \leftarrow I \in \mathbb{R}^{N \times H_0 \times W_0 \times 3}$ \\
    $I'_{\text{grid}} = \text{DELIGHT}(I_{\text{grid}})$ \\
    $I_{d} \in \mathbb{R}^{N \times H_0 \times W_0 \times 3} \leftarrow I'_{\text{grid}} \in \mathbb{R}^{(N \times H_0) \times W_0 \times 3}$
}

\SetKwBlock{SR}{2. Super Resolution Color Image}{}
\SR{
    $I_{sr} = \{\text{SR}(I_{d, i}) \mid i = 1, \ldots, N \}$ \\
    $I_{sr} \in \mathbb{R}^{N \times H \times W \times 3} \quad \text{where} \quad H = H_0 \times \text{upscale\_h}, \, W = W_0 \times \text{upscale\_w}$
}

\SetKwBlock{ComputeGeoBuffer}{3. Mesh Geometry Buffer Rendering}{}
\ComputeGeoBuffer{
    $M\in\{0,1\}^{N\times H\times W}$: per‐pixel visibility mask;\\
    $D\in[0,1]^{N\times H\times W}$: normalized depth map;\\
    $N\in\mathbb{R}^{N\times H\times W\times 3}$: view‐space normals;\\
    $U\in[0,1]^{N\times H\times W\times 2}$: per‐pixel UV;\\
}

\SetKwBlock{BackProj}{4. Back-Projection per View then Fusion}{}
\BackProj{
    $\mathbf{v}=[0,0,1]^T$ \COMMENT{Camera view direction} \\
    Initialize $T = 0$ and $C = 0$ \COMMENT{Texture and confidence map} \\
    \For{each view $i=1$ to $N$}{
        $C_i \leftarrow \max(0, N_i \cdot \mathbf{v})$ \\
        $C_i[C_i < \cos(\theta)] \leftarrow 0$ \COMMENT{Exclude large angles} \\
        $E_i \leftarrow \mathrm{Canny}(D_i)$ \\
        $M_i' \leftarrow M_i \land (E_i < 0.5)$ \COMMENT{Exclude edge} \\
        $C_i[M_i' = 0] \leftarrow 0$ \\
        $C_i \leftarrow C_i \times W_i$ \\
         \For{each valid pixel $p$ where $M_i'[p] = 1$}{
            $(u,v) \leftarrow U_i[p] \times [W_{\mathrm{tex}}, H_{\mathrm{tex}}]$ \COMMENT{Map UV to texture space} \\
            Scatter $I_i[p] \times C_i[p]$ to $T(u, v)$ \COMMENT{Scatter color with confidence weight} \\
            Scatter $C_i[p]$ to $C(u, v)$ \COMMENT{Scatter confidence map} \\
        }
    }
    $T = \frac{T}{C + \epsilon}$ \COMMENT{Texture fusion by confidence}
}
\Return{$T$}
\end{algorithm}

\subsection{Text-to-3D Object Generation}
\label{sec:robo_asset_gen_text}
\begin{figure}[!htbp]
\centering
\includegraphics[width=\linewidth]{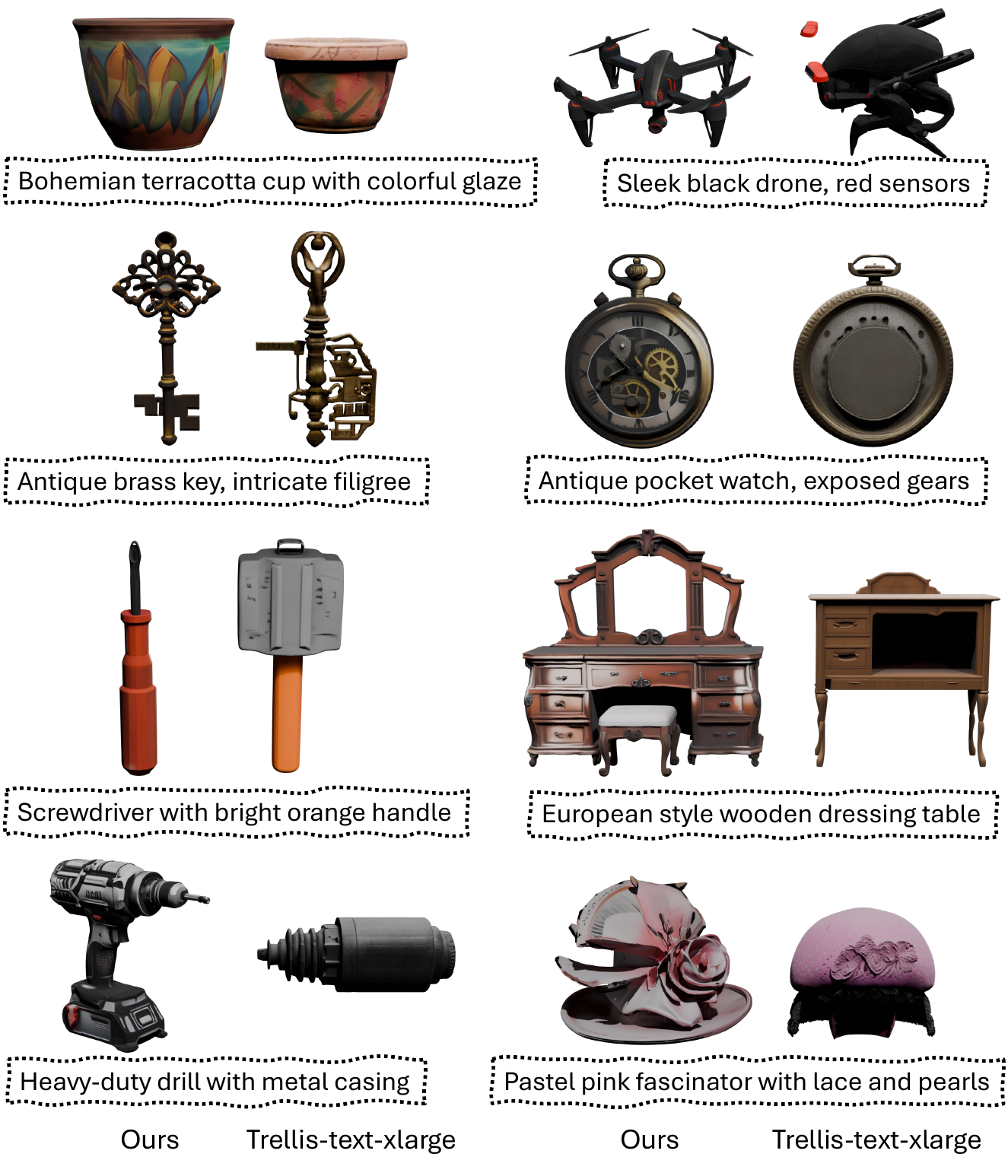}
\vspace*{-7mm}
\caption{Qualitative comparison of Text-to-3D result. The left column shows results generated by our method, while the right column shows results from TRELLIS-text-xlarge. Our method produces significantly higher-quality outputs that better align with the input text descriptions.}
\vspace*{-3mm}
\label{fig:text_compare}
\end{figure}

\begin{figure}[!htbp]
\centering
\includegraphics[width=\linewidth]{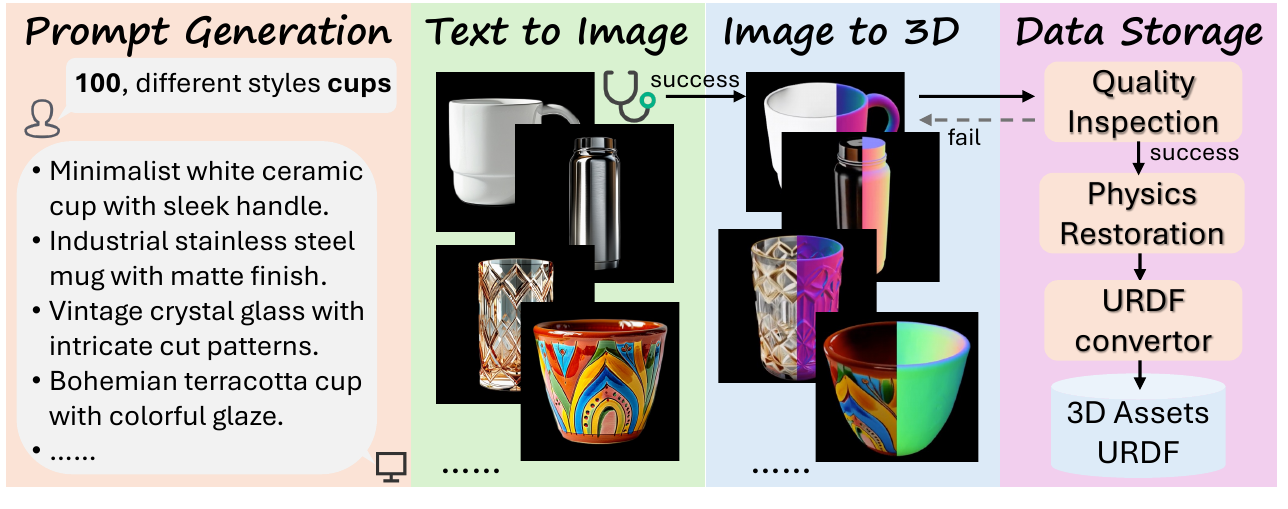}
\vspace*{-7mm}
\caption{\textit{{\mymethod}} Text-to-3D module for large-scale 3D asset generation. A prompt generator decomposes user requirements into prompts targeting different asset styles. The pipeline proceeds through text-to-image and image-to-3D stages, each equipped with automatic quality inspection and retry mechanisms. The final URDF asset with complete geometry, realistic scale, and physical properties, is persistently stored.}
\vspace*{-3mm}
\label{fig:textto3d}
\end{figure}

\paragraph{Method Overview}
The Text-to-3D module is designed for highly controllable generation of 3D object assets with diverse geometry and textures. To achieve this, the text-to-3D task is decomposed into two stages: text-to-image and image-to-3D. This decoupling brings several advantages. In large-scale asset production, it enables early-stage automatic quality inspection, allowing the system to filter out samples that fail foreground segmentation check or contain semantics inconsistent with the text description before committing computational resources to 3D generation. More importantly, this modular design improves iteration flexibility and reduces maintenance costs. It also allows the pipeline to fully benefit from ongoing advancements in the text-to-image and image-to-3D communities, supporting continuous improvement in controllability, scalability, and asset generation quality. Specifically, Kolors\cite{kolors} is used as our text-to-image generation model, as it supports high-quality image generation from both Chinese and English prompts. For the image-to-3D stage, we maintain a single unified service, \textit{{\mymethod}} Image-to-3D, to streamline system complexity. 
As shown in Figure~\ref{fig:text_compare}, compared to Trellis-text-xlarge\cite{xiang2024structured}, our two-stage design offers improved controllability and generation quality, while significantly reducing the maintenance cost associated with end-to-end text-to-3D models. The large-scale asset generation workflow for text-to-3D is illustrated in Figure~\ref{fig:textto3d}.

\paragraph{Evaluation of Automated Quality Inspection}
We evaluate the efficiency of the automated quality inspection module in large-scale 3d asset generation. We construct the automated quality inspection pipeline based on \textit{AestheticChecker}, \textit{ImageSegChecker}, and \textit{MeshGeoChecker}, as introduced in Section~\ref{sec:robo_asset_gen_img}. During evaluation, a generated asset is considered \textit{usable} if it satisfies the following criteria: geometric and textural consistency with the input text description, geometric completeness, texture richness, and compatibility with simulation engines. Otherwise, it is classified as \textit{unusable}.
We define \textbf{precision} as the proportion of assets identified as unusable by the automated checkers that are indeed unusable, and \textbf{recall} as the proportion of all truly unusable assets that are correctly flagged by the checkers.
We generated 150 cup 3d assets and manually annotated them. Among these, 107 were labeled as usable and 43 as unusable. The automated quality inspection achieved a precision of \textbf{68.7\%} and a recall of \textbf{76.7\%}. While these metrics are not yet above 90\%, the current system substantially reduces the manual effort required for asset screening. Moreover, we expect that this pipeline will continue to improve as multi-modal large models advance, further enhancing automated quality assessment in the future.

\subsection{Articulated Object Generation}
\label{sec:articulated_gen}

Articulated objects such as cabinets, drawers and appliances are common in real world environments. Modeling these objects accurately requires not only capturing their geometric structure but also understanding their motion behavior and part connectivity. This capability is fundamental for tasks in virtual simulation, robotics, and interactive environments~\cite{gadre2021act,mo2021where2act,qian2023understanding}.

\paragraph{Method Overview} We use DIPO~\cite{wu2025dipodualstateimagescontrolled}, a controllable generation framework that constructs articulated 3D objects from a dual-state image pair. One image shows the object in a resting state, and the other shows it in an articulated state. This dual-state input format encodes both structural and kinematic information, enabling the model to better resolve motion ambiguity and predict joint behavior. The generation process is based on a diffusion transformer that integrates these two images using a specialized dual-state injection module at each layer. DIPO also includes a chain-of-thought based Graph Reasoner that infers connectivity relationship between each part. The resulting articulation graph is used as an attention prior to enhance generation consistency and plausibility.

\begin{figure}
    \centering
    \includegraphics[width=\linewidth]{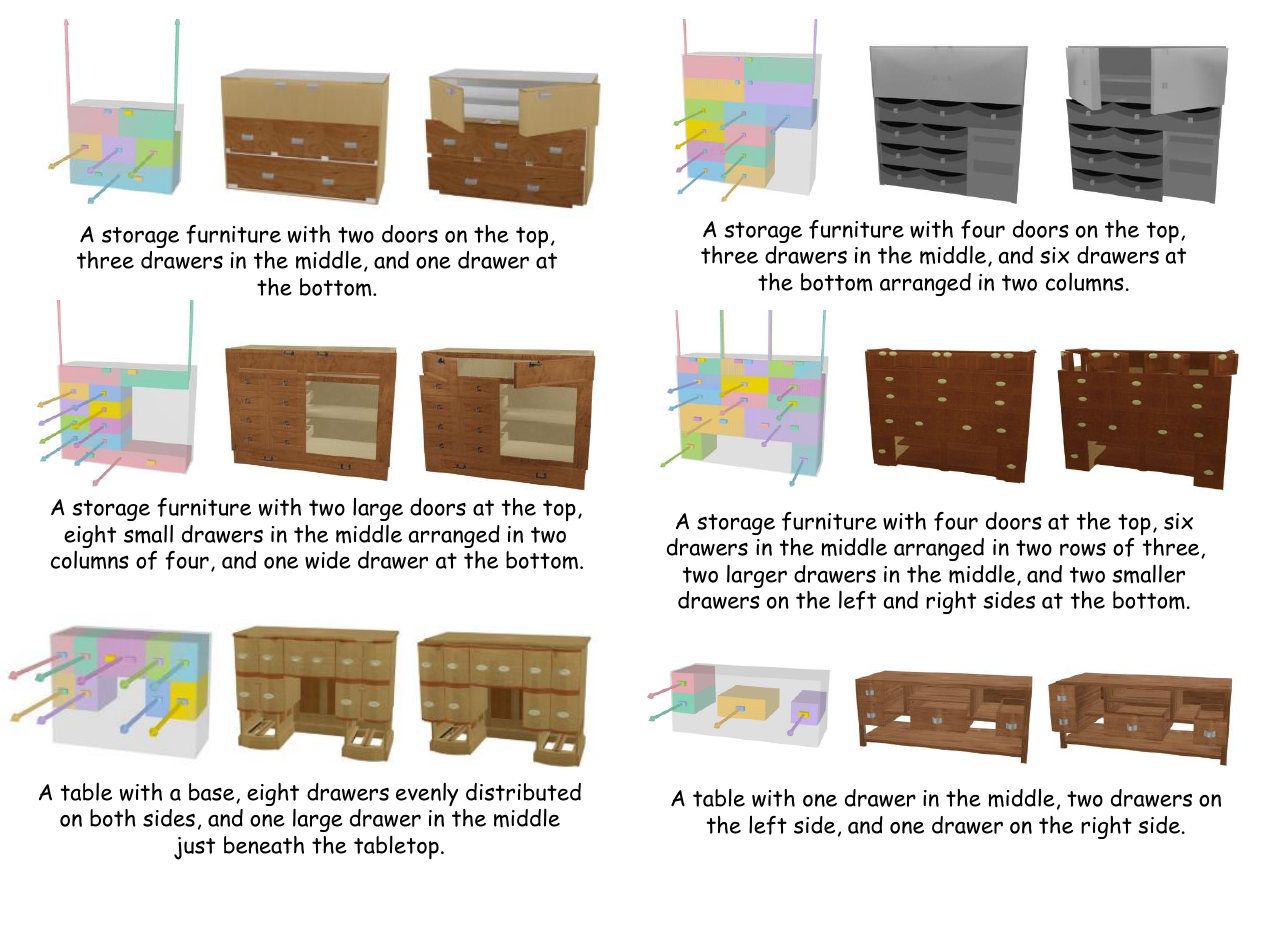}
    \vspace*{-10mm}
    \caption{Visual examples of articulated objects constructed by the LLM agent based workflow.}
    \vspace*{-3mm}
    \label{fig:pmx}
\end{figure}

\paragraph{Automatic Articulated Object Data Augmentation}
Beyond model design, to improve generalization on complex articulated object generation, we use an automatic data augmentation pipeline to synthesize articulated object layouts from natural language prompts using grid-based spatial reasoning and part retrieval from existing 3D datasets. The resulting PM-X dataset comprises 600 structurally diverse articulated objects, each annotated with rendered images and physical properties. Figure~\ref{fig:pmx} shows representative examples from the PM-X dataset.

\paragraph{Qualitative comparison} Figures~\ref{fig:results} illustrates representative results of generated objects with real-world image inputs. Explore more demos of dynamic articulated object on our project page.

\subsection{Texture Generation}
\label{sec:robo_asset_gen_edit}

\begin{figure*}[!h]
\centering
\includegraphics[width=\linewidth]{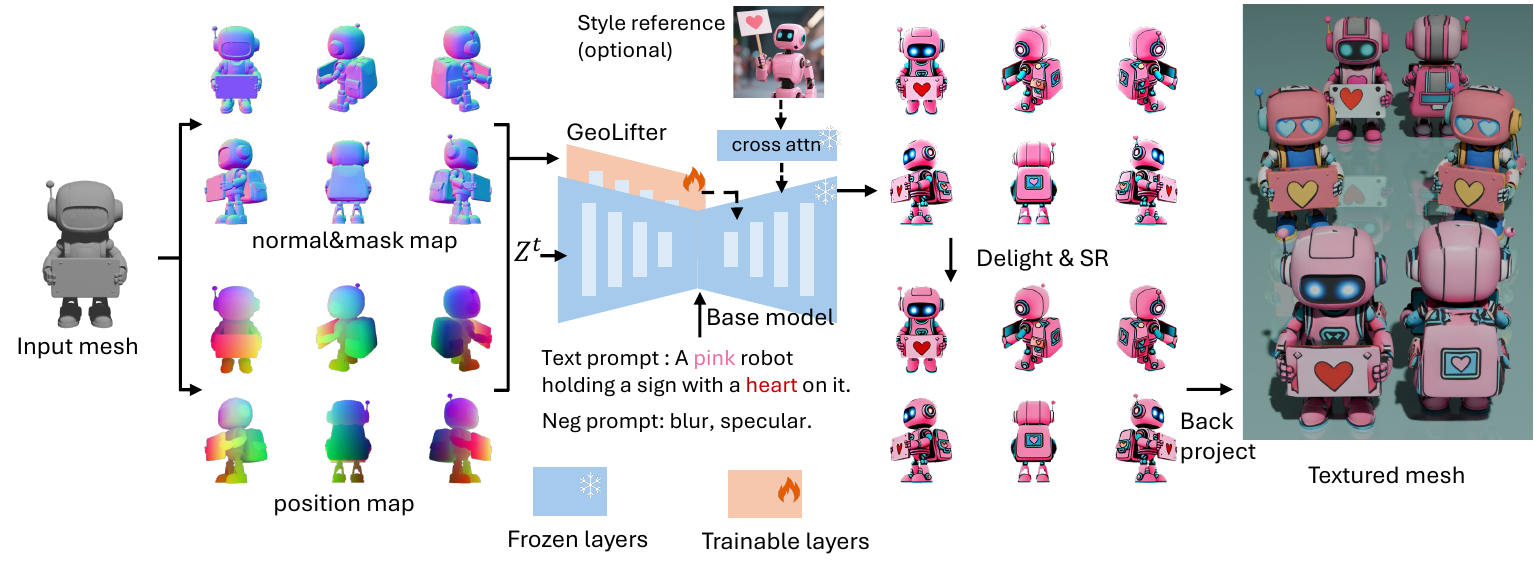}
\vspace*{-8mm}
\caption{Overview of \textit{{\mymethod}} Texture Generation Module. Given a mesh and a text prompt, the module generates six-view consistent textures with controllable styles via text, reference image, or both. Geometry-aware conditions (normals, positions, masks) are extracted and injected into a diffusion model via the \textit{GeoLifter} module. The outputs are refined with illumination removal and super-resolution, then back-projected to the mesh as described in Algorithm~\ref{alg:back_project}.
}
\vspace*{-3mm}
\label{fig:edit_photo}
\end{figure*}

\paragraph{Method Overview} The texture generation module is designed to perform multi-style texture generation and editing for 3D object assets. Given a 3D mesh as input, it outputs a textured 3D mesh with generated visual appearance. Instead of training a multi-view diffusion model from scratch, we design a plug-gable and extensible module that leverages existing 2D text-to-image foundation models and extends their capabilities into the 3D domain. Our approach enables the generation of diverse and high-quality textures that are geometrically consistent across views. This design paradigm allows us to capitalize on continuous improvements in community foundation models, enabling cost-effective and scalable generation of view-consistent textures with minimal retraining effort.
\paragraph{Model Design} We develop a model named \textit{GeoLifter}, a module that extends the capabilities of foundation text-to-image diffusion models to multi-view generation with geometric consistency. \textit{GeoLifter} injects geometric control into the base diffusion model through cross-attention mechanisms, enabling view-consistent texture generation conditioned on 3D geometry. We adopt Kolors text-to-image\cite{kolors} as the base diffusion model. In contrast to approaches such as ControlNet\cite{zhang2023adding}, which duplicate and train a separate encoder branch of the base model’s U-Net. \textit{GeoLifter} remains lightweight and highly extensible. Its parameter size does not grow with the depth of the base model, making it more efficient and easier to integrate with evolving diffusion architectures.

Given an input mesh, we render normal maps, position maps and binary masks from six predefined camera views (elevations $\in \{20^\circ, -10^\circ\}$, azimuths $\in \{0^\circ, 60^\circ, 120^\circ, 180^\circ, 240^\circ, 300^\circ\}$). For each view, the normal and position maps are rendered in image space from camera view, and concatenated along the spatial (height and width) dimensions within each attribute. The different attribute types (normal, position, mask) are then concatenated along the channel dimension to form the geometric condition input $G \in \mathbb{R}^{H \times W \times 7}$. The normal map encodes surface normals interpolated per vertex and projected to the image plane. The position map stores the XYZ coordinates (in object space) of visible vertices. The mask is a binary segmentation map. The geometric condition $G$ is then implicitly encoded into a feature embedding, which is progressively injected into the denoising process of the diffusion model via cross-attention, leveraging zero convolution to ensure minimal interference with the base model decoder at the start of training.

The text prompt supports both positive and negative prompts, and accepts multi-lingual input, including both Chinese and English descriptions, to specify the desired texture style and appearance. In addition to textual prompts, users may optionally provide an RGB image as a reference style, which serves as a complementary control signal to the language input. Users can provide a text prompt only, a reference image only, or both simultaneously. This design enables highly controllable and expressive texture generation by jointly leveraging semantic guidance and visual style cues.

We observe that \textit{GeoLifter}, with its lightweight geometric conditioning design, effectively preserves the texture generation capability of the underlying foundation model, while significantly improving spatial and geometric consistency across views. Following the multi-view texture generation, we apply illumination removal and super-resolution techniques and project the refined textures back into 3D space to obtain the final textured mesh, equipped with a high-resolution 2K UV map as described in Algorithm~\ref{alg:back_project}.

\begin{figure}
\centering
\includegraphics[width=\linewidth]{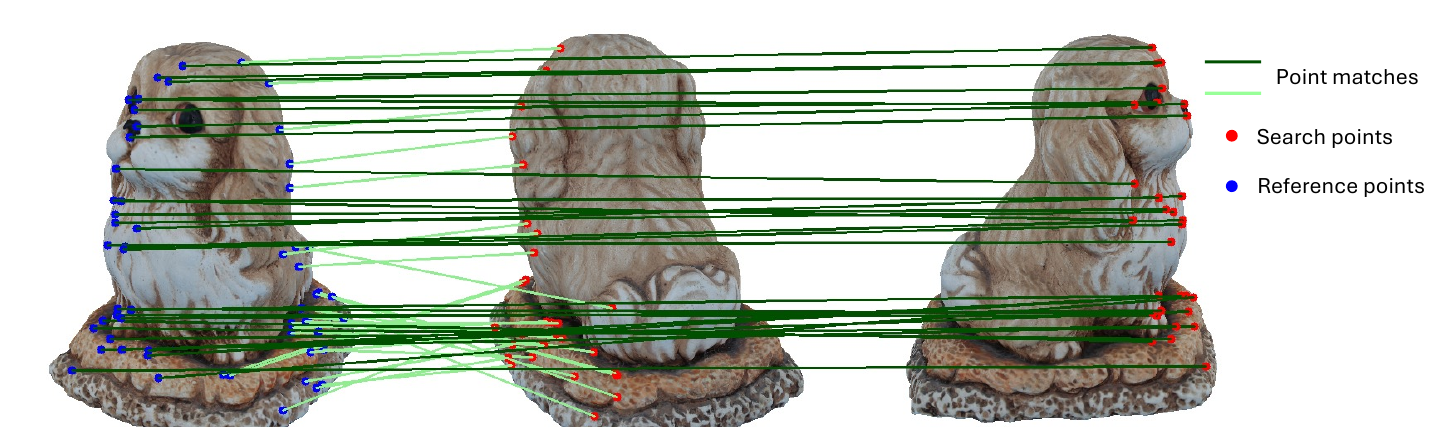}
\vspace*{-7mm}
\caption{Blue dots: reference points; red dots: projected correspondences in other views; green lines: matches. The spatial loss is applied to let the latent features of matched points closer, enhancing cross-view alignment.}
\vspace*{-3mm}
\label{fig:edit_loss}
\end{figure}

\paragraph{Loss Design} On top of the original loss~\ref{eq:ldm_loss} used in latent diffusion models\cite{ldm}, we introduce spatial loss~\ref{eq:spatial_loss} as a geometric consistency constraint in the latent space, where $B$ is the batch size, $\mathbf{r}_b$ and $\mathbf{s}_b$ are the reference and search point sets for the $b$-th sample(visualized in Figure~\ref{fig:edit_loss}). $f_b(\cdot) \in \mathbb{R}^{C \times N_b}$ denotes the extracted feature vectors at the corresponding coordinates. Standard element-wise smooth L1 loss is applied to encourages the latent features of pixels corresponding to the same 3D point (projected across multiple views) to remain close in feature space, thereby enhancing cross-view coherence. The final loss~\ref{eq:total_loss} is obtained by adding $\mathcal{L}_{\text{LDM}}$ and $\mathcal{L}_{\text{spatial}}$, $\lambda_{ldm}$ and $\lambda_{spatial}$ is set to 1 and 0.02 respectively.

\begin{equation}
\label{eq:ldm_loss}
\mathcal{L}_{\text{LDM}} = \mathbb{E}_{\mathbf{x}, \boldsymbol{\epsilon}, t} \left[ \left\| \boldsymbol{\epsilon} - \boldsymbol{\epsilon}_\theta\left(\mathbf{z}_t, t, \mathbf{c} \right) \right\|^2 \right]
\end{equation}

\begin{equation}
\label{eq:spatial_loss}
\mathcal{L}_{\text{spatial}} = \frac{1}{B} \sum_{b=1}^{B} \mathbf{1}_{\{|\mathbf{r}_b| > 0 \land |\mathbf{s}_b| > 0\}} \cdot \text{SmoothL1} \left( f_b(\mathbf{r}_b), f_b(\mathbf{s}_b) \right)
\end{equation}

\begin{equation}
\label{eq:total_loss}
\mathcal{L} = \lambda_{\text{LDM}} \mathcal{L}_{\text{LDM}} + \lambda_{\text{spatial}} \mathcal{L}_{\text{spatial}}
\end{equation}

\begin{figure*}
\centering
\includegraphics[width=0.81\linewidth]{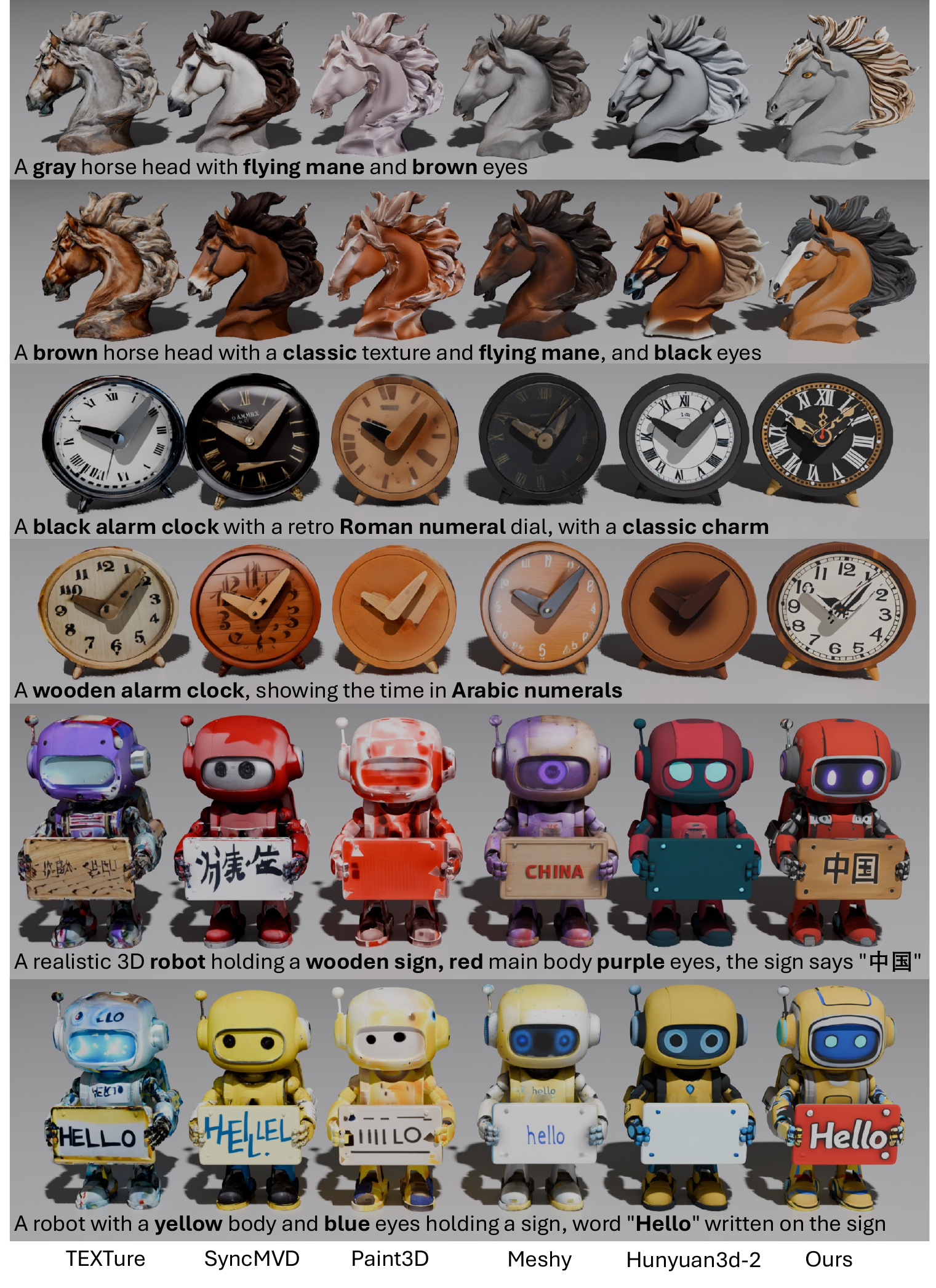}
\vspace*{-5mm}
\caption{\textit{{\mymethod}} texture generation module effectively adheres to text descriptions, generating high-quality textures with strong spatial and geometric consistency. It also demonstrates robust control over text generation on textures, accurately rendering common Chinese and English text.}
\vspace*{-3mm}
\label{fig:edit_compare}
\end{figure*}

\paragraph{Qualitative Comparison} 
We conduct a qualitative comparison between our method and several state-of-the-art texture generation methods, including TEXTure\cite{texture}, SyncMVD\cite{liu2023text}, Paint3D\cite{zeng2024paint3d}, Meshy\cite{meshyai}, and Hunyuan3d-2\cite{zhao2025hunyuan3d20scalingdiffusion}. As shown in Figure~\ref{fig:edit_compare}, our method consistently produces higher-quality 3D textures, with superior geometric consistency across views. Furthermore, our method uniquely supports specified text to be directly generated on 3D surfaces.

% \clearpage
\newpage

\subsection{3D Scene Generation}
\label{sec:robo_scene_gen}

\begin{figure*}[t]
\centering
\includegraphics[width=\linewidth]{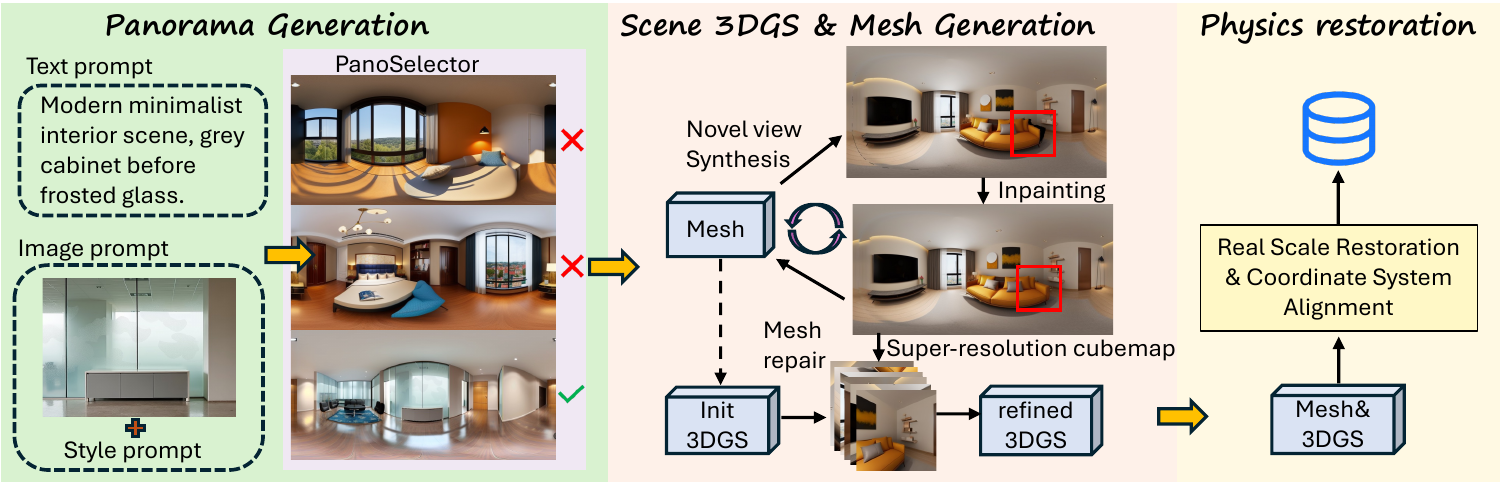}
\vspace*{-7mm}
\caption{Overview of \textit{{\mymethod}} 3D Scene Generation. A panorama is generated from a text prompt or input image, guided by a style prompt. After quality assessment via vlm-based selector, a refined mesh and 3DGS\cite{kerbl20233dgaussiansplattingrealtime} are generated by panorama projection, inpainting, and repair. Super-resolution is used to enhance 3DGS appearance details, followed by real scale and alignment adjustments.}
\vspace*{-3mm}
\label{fig:scene_pipeline}
\end{figure*}

\paragraph{Method Overview}
Beyond 3D object asset generation, scene diversity as background context plays an equally critical role. We develop a scalable and efficient framework for 3D scene generation. The system follows a modular pipeline that transforms multi-modal inputs into panoramic images, which are then used to generate 3D scenes with consistent real-world scale. The framework consists of three main stages: (1) panoramic image generation, (2) 3D scene generation in 3DGS\cite{kerbl20233dgaussiansplattingrealtime} and mesh representation from panorama, and (3)scale alignment and standardized output, as illustrated in Figure~\ref{fig:scene_pipeline}.

\begin{figure}
\centering
\includegraphics[width=\linewidth]{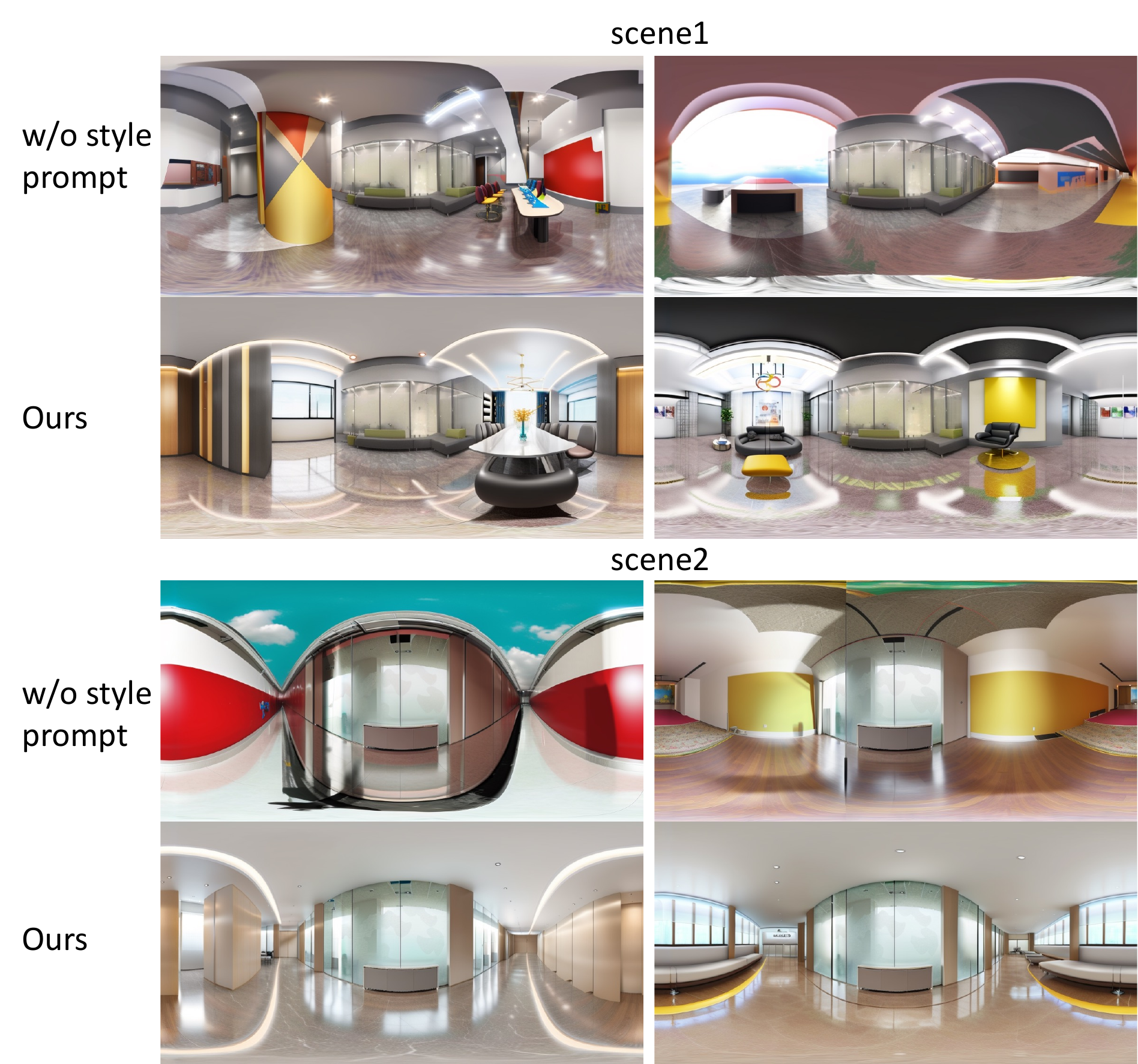}
\vspace*{-7mm}\caption{Qualitative Comparison With and Without Style Prompts. “w/o style prompt” lacks explicit style guidance, while “Ours” uses style-aware prompting, yielding more coherent textures and better stylistic alignment across scenes.}
\vspace*{-3mm}
\label{fig:wo_style_prompt}
\end{figure}

\paragraph{Panoramic Image Generation}
Our method supports both text- and image-based input modalities, or a combination of both, enabling flexible and efficient generation of high-quality panoramic images. For text-driven generation, user-provided scene descriptions are translated into panoramic views using Diffusion360 model\cite{feng2023diffusion360seamless360degree}, which have demonstrated strong performance in this task. For image-driven generation, we employ Qwen~\cite{qwen} to extract semantic descriptions from the input image. The image and its corresponding textual description are then jointly processed by the panorama generation model\cite{feng2023diffusion360seamless360degree} to generate semantically aligned panoramas(see Figure~\ref{fig:wo_style_prompt}). To ensure quality and reliability, we introduce the \textit{PanoSelector} module, which is built upon Qwen~\cite{qwen}, automatically evaluates and filters the generated panoramas based on structural quality metrics, such as floor and wall consistency. This guarantees that only high-quality outputs are forwarded to the geometry generation stage.

\begin{figure}[ht]
\centering
\includegraphics[width=\linewidth]{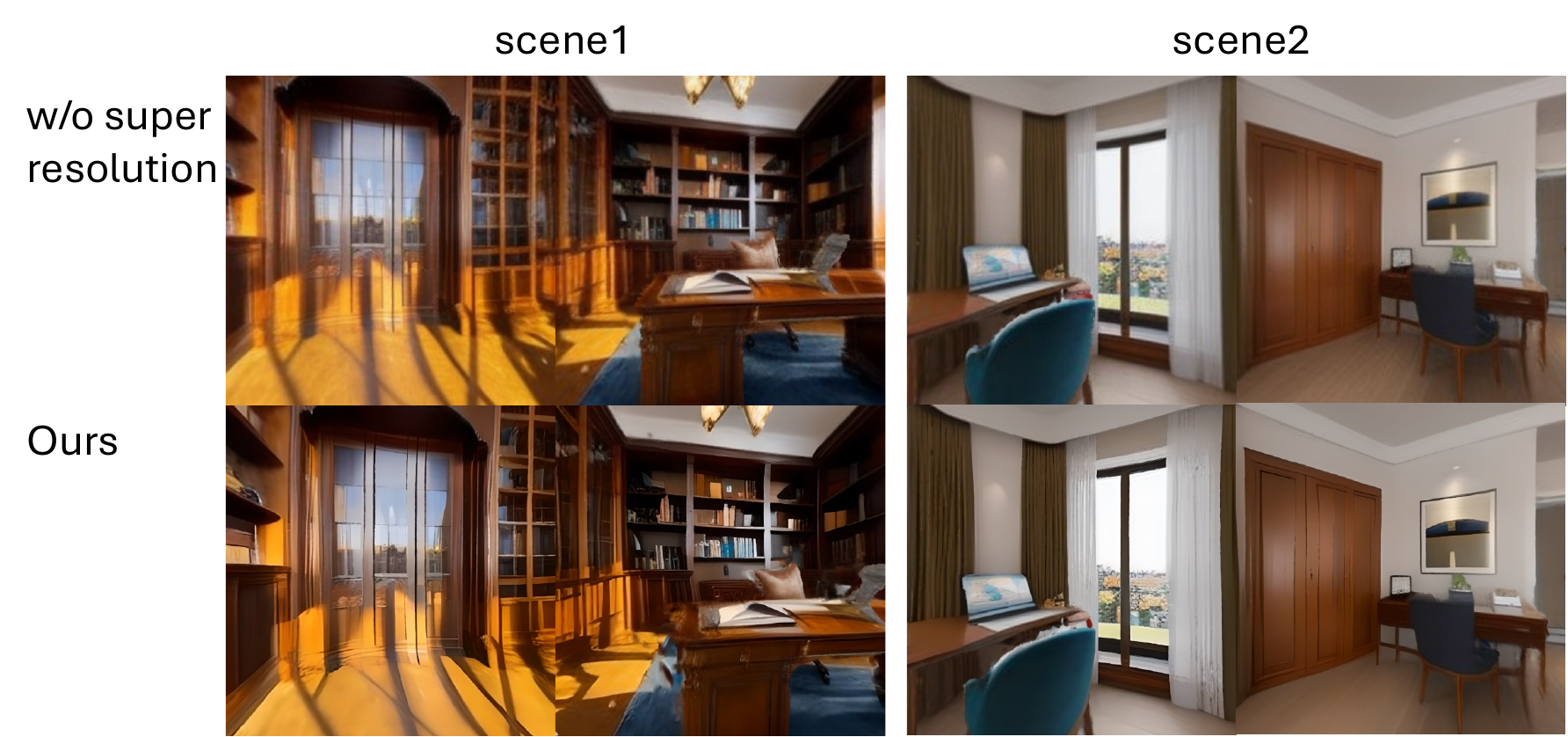}
\vspace*{-7mm}
\caption{Qualitative comparison with and without super-resolution. The generated 3D scene show sharper and high-frequency detailed with super-resolution. Zoom in for details.
}
\vspace*{-3mm}
\label{fig:wo_super_resolution}
\end{figure}

\begin{figure}
\centering
\includegraphics[width=\linewidth]{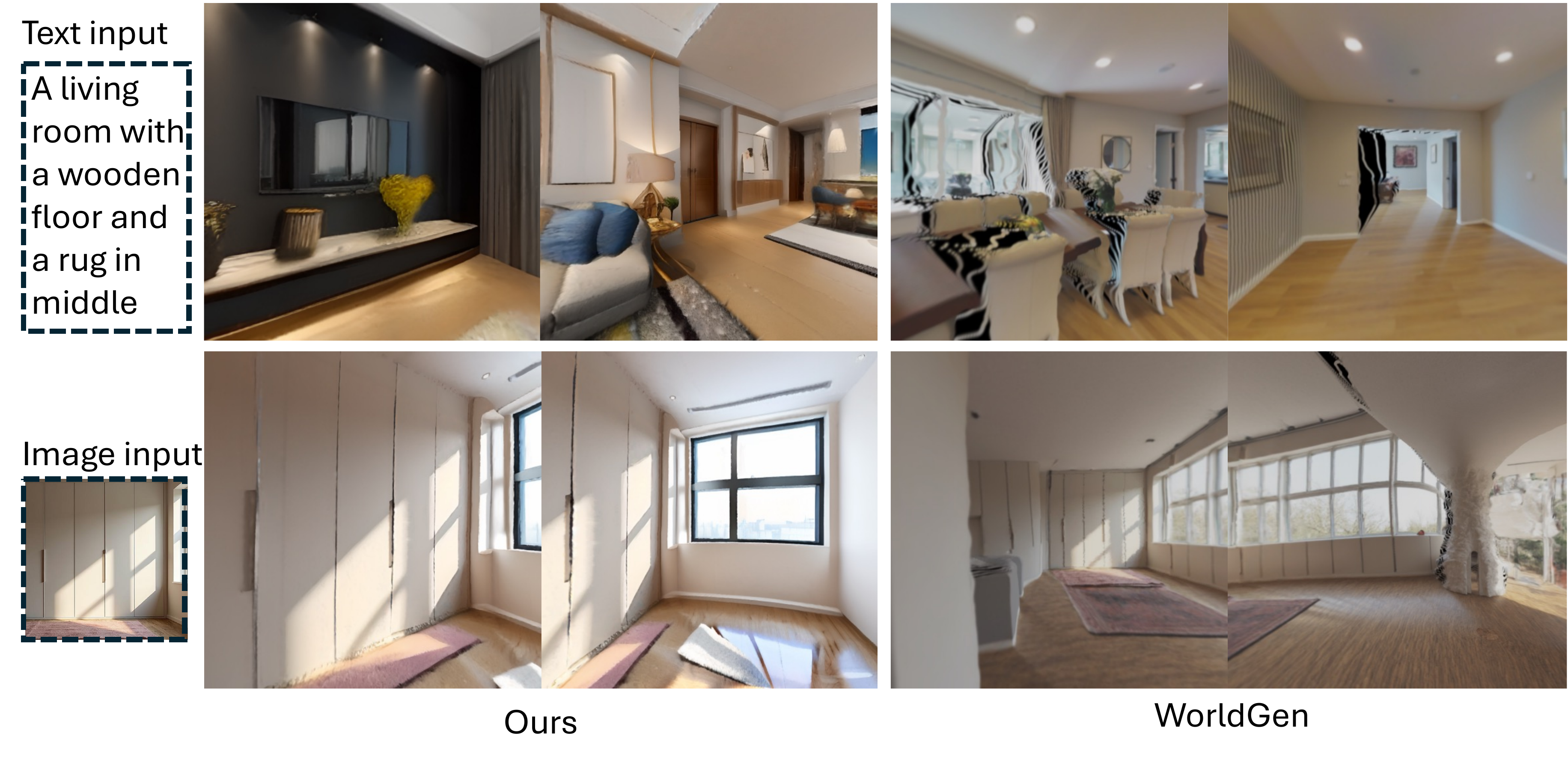}
\vspace*{-7mm}
\caption{Qualitative Comparison with WorldGen\cite{worldgen2025ziyangxie} Our method produces more detailed textures and more complete geometry than WorldGen, under both text and image input settings.}
\vspace*{-3mm}
\label{fig:scene_compare}
\end{figure}

\paragraph{Scene 3D Representation Generation}
After obtaining high-quality panoramas, the system proceeds to generate the corresponding 3D representation in 3DGS and mesh based on Pano2Room\cite{Pu_2024}. An initial mesh is generated from the panoramic input and further refined through mesh optimization to improve both geometric accuracy and reconstruct-ability. The optimized mesh is then converted into a 3DGS representation. To enhance visual fidelity, views rendered from the optimized mesh are converted into cubemaps and passed through the super-resolution model\cite{wang2021realesrgan}. The super-resolved images are then used to further refine the initial 3DGS, effectively enhancing the detail quality of the final 3DGS output as in Figure~\ref{fig:wo_super_resolution}. We show generated 3D scene quality comparison with worldgen\cite{worldgen2025ziyangxie} in Figure~\ref{fig:scene_compare}.

\paragraph{Physics Restoration}
To produce realistic and metrically consistent 3D scenes, the system performs absolute scale estimation by predicting real-world dimensions such as building height from the input panoramas and their semantic descriptions. A dedicated scale estimation module, built upon the Qwen model~\cite{qwen}, infers these scale factors to enable lossless rescaling of both the mesh and 3DGS\cite{kerbl20233dgaussiansplattingrealtime} representations. Additionally, the coordinate system is re-centered to the floor plane of the scene, with the axes aligned according to either the camera direction from the input image or the orientation implied by the textual description. The resulting output is a scale-aligned, high-fidelity 3D scene asset, ready for downstream use in virtual and augmented reality and robotics.

\section{Application}
\label{sec:application}

\begin{figure*}
\centering
\includegraphics[width=\linewidth]  {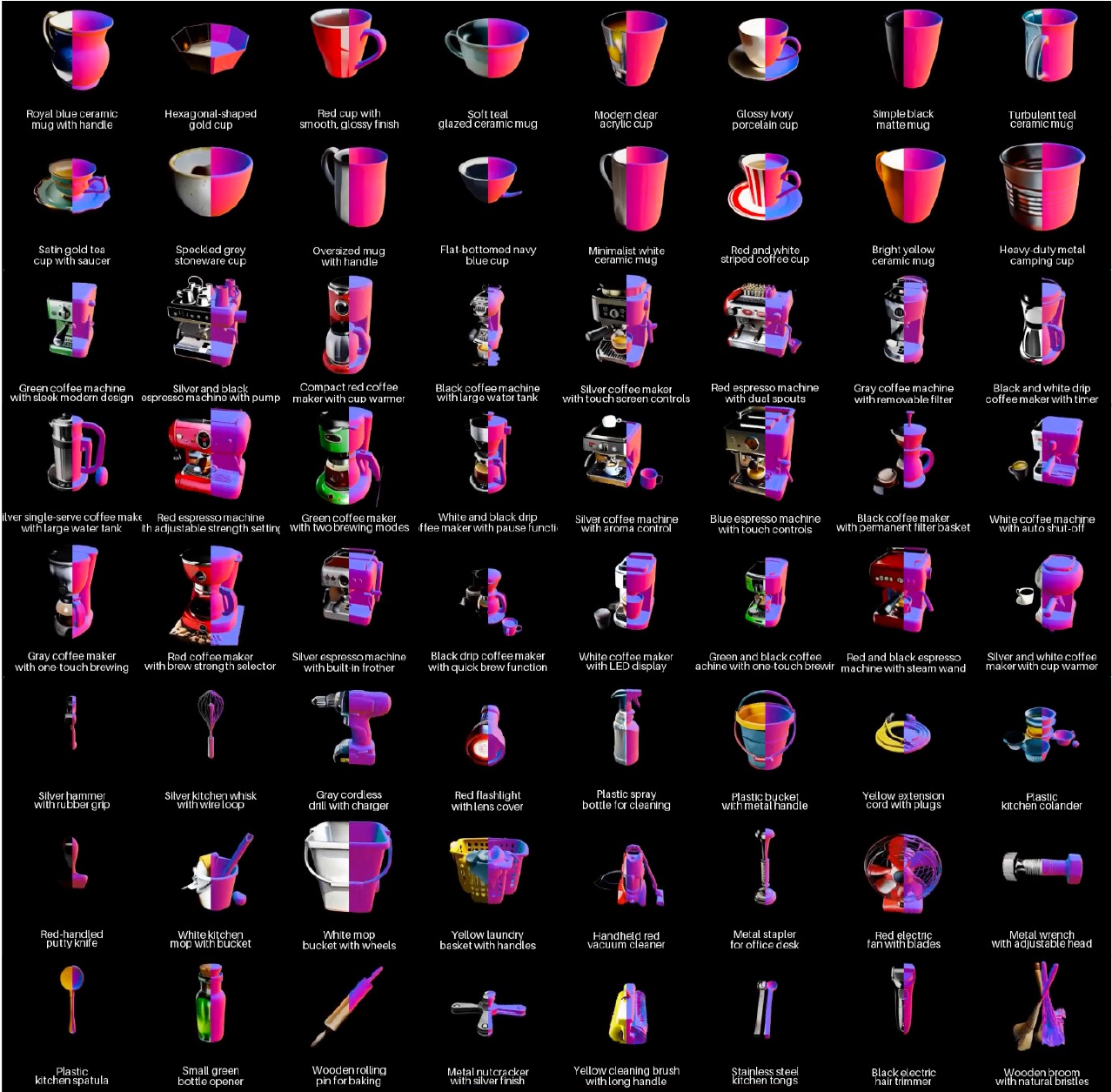}
\vspace*{-7mm}
\caption{\textit{{\mymethod}} Image-to-3D: large-scale and diverse 3D object asset generation.}
\vspace*{-3mm}
\label{fig:app1}
\end{figure*}

\paragraph{Large-scale 3D Asset Generation}
Figure~\ref{fig:app1} showcases the capability of the \textit{\mymethod} Text-to-3D module to generate large-scale 3D assets for embodied intelligence tasks, producing watertight and stylistically diverse meshes aligned with textual descriptions. This capability enables low-cost augmentation of interactive 3D assets for simulation and downstream training and evaluation.

\begin{figure*}
\centering
\includegraphics[width=\linewidth]{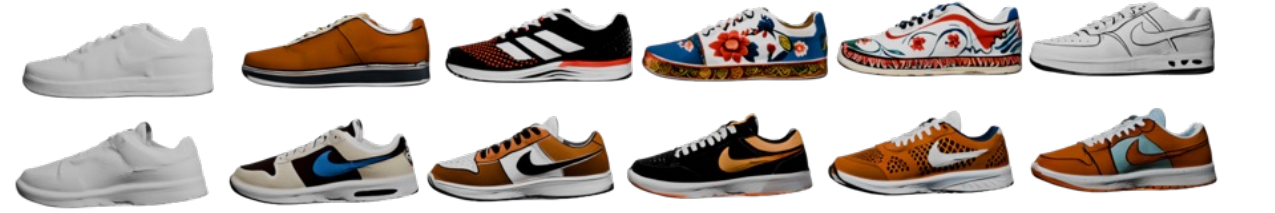}
\vspace*{-7mm}
\caption{
\textit{{\mymethod}} texture generation module enables rich and flexible visual texture editing.}
\vspace*{-3mm}
\label{fig:shoes}
\end{figure*}

\paragraph{Visual Appearance Editing of 3D Mesh}
Figure~\ref{fig:shoes} demonstrates the capability of the \textit{\mymethod} texture generation module to generate and edit photorealistic textures with rich visual details. The edited 3D assets can be used for training data augmentation, enhancing model generalization in visual appearance understanding.

\begin{figure*}
\centering
\includegraphics[width=0.75\linewidth]
{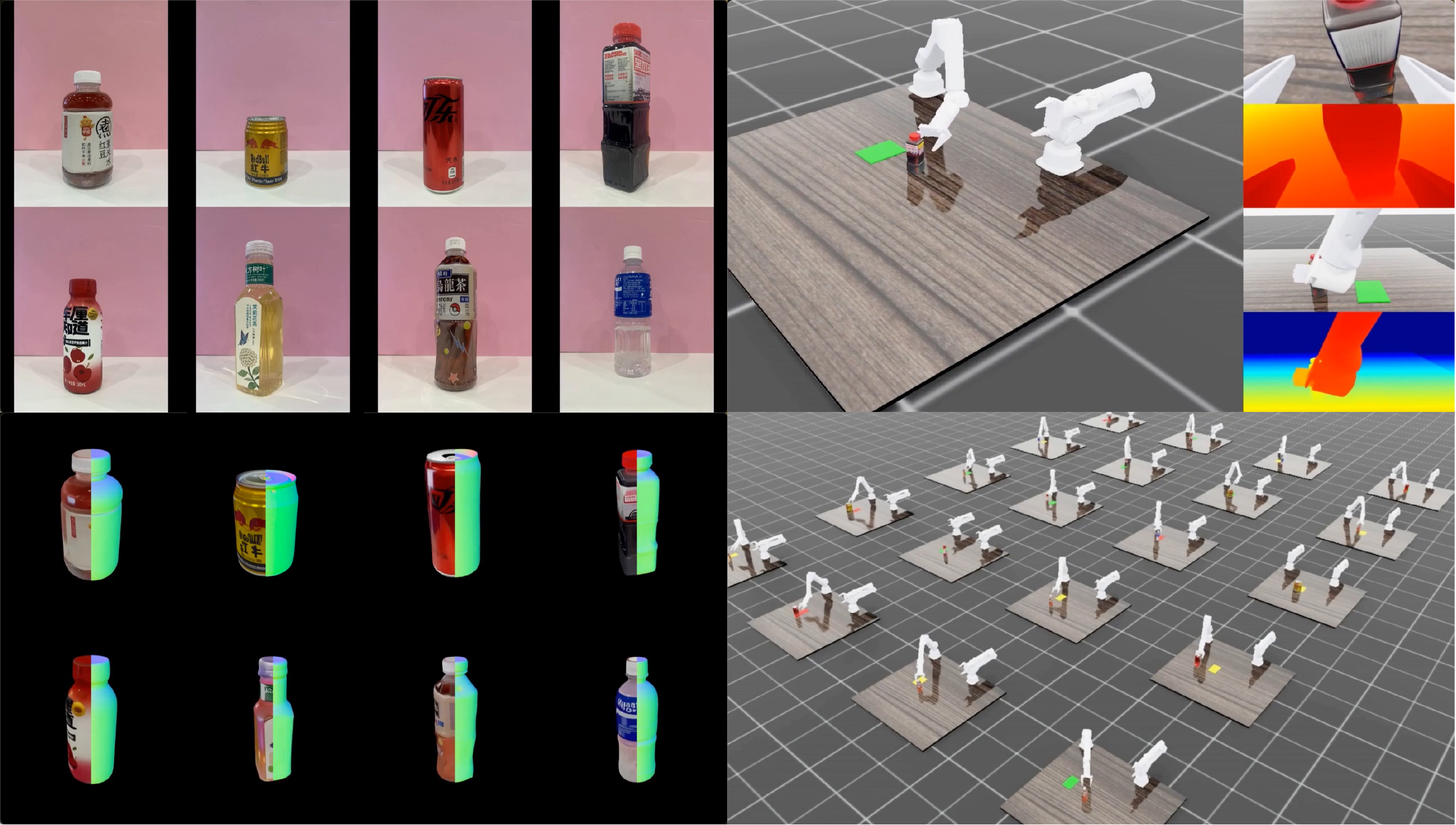}
\vspace*{-2mm}
\caption{\textit{{\mymethod}} Image-to-3D: Real-to-sim closed-loop simulation evaluation of a grasping model in Isaac Lab environment \cite{mittal2023orbit}.}
\vspace*{-3mm}
\label{fig:app3}
\end{figure*}

\begin{figure*}
\centering
\includegraphics[width=0.75\linewidth]{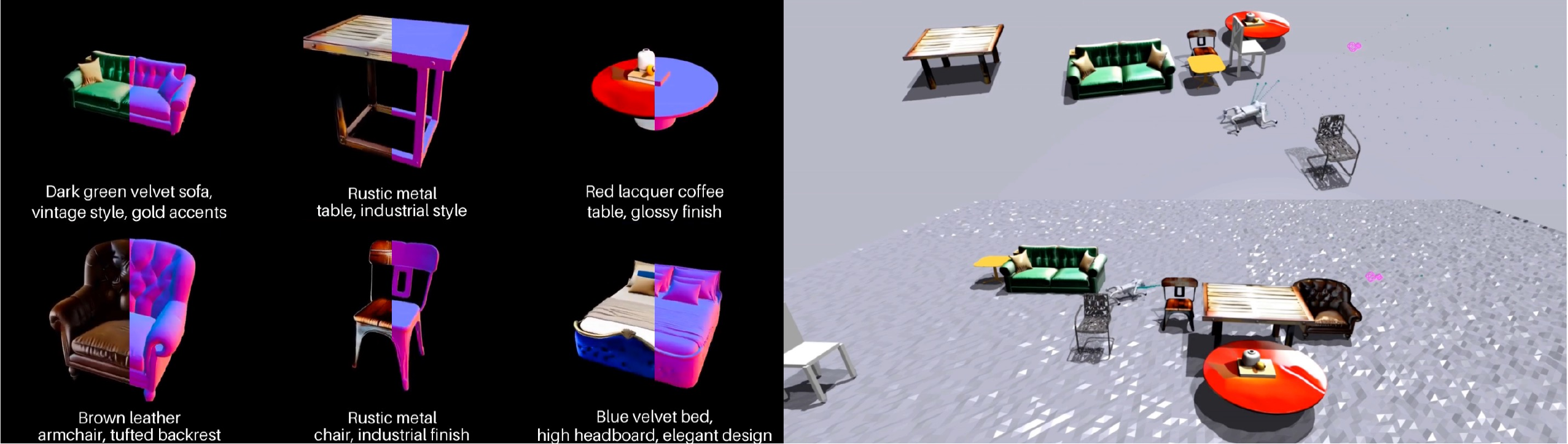}
\vspace*{-2mm}
\caption{\textit{{\mymethod}} Text-to-3D: Real-to-sim object transfer and quadruped navigation with obstacle avoidance in OpenAI Gym\cite{openai_gym}.}
\vspace*{-2mm}
\label{fig:app4}
\end{figure*}

\paragraph{Real-to-Sim: Digital Twins creation} 
The real-to-sim capabilities of \textit{{\mymethod}} Image-to-3D module is illustrated in Figure~\ref{fig:app3}, with the generated 3D assets evaluated via closed loop simulation in the Isaac Lab\cite{mittal2023orbit} environment.
Figure~\ref{fig:app4} shows how assets generated by \textit{{\mymethod}} Text-to-3D can be used in navigation and obstacle avoidance tasks within the OpenAI Gym\cite{openai_gym} simulation framework.

\begin{figure*}
\centering
\includegraphics[width=\linewidth]{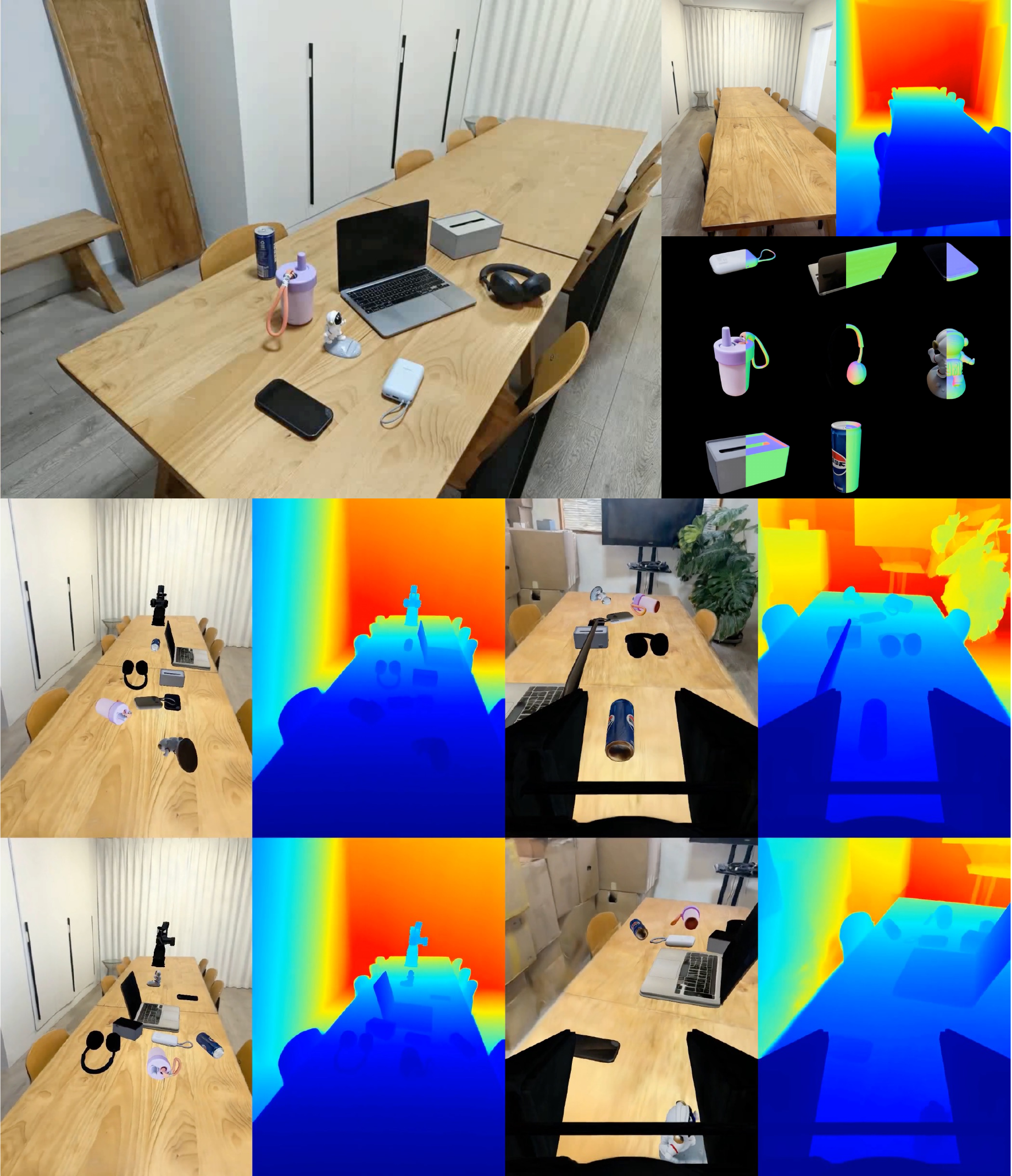}
\vspace*{-7mm}
\caption{\textit{{\mymethod}} Image-to-3D: Digital twin creation and simulation in \textit{RoboSplatter} and MuJoCo\cite{todorov2012mujoco}.}
\vspace*{-3mm}
\label{fig:app2}
\end{figure*}

\paragraph{\textit{\textbf{RoboSplatter}}: Integration of 3DGS Rendering into Physical Simulation}
Existing simulators are typically built upon traditional OpenGL-based rendering techniques, which involve complex environment modeling, lighting setup, and ray-based rendering calculations. These approaches often suffer from high computational cost and limited photorealism. With the rapid advancement of 3DGS\cite{kerbl20233dgaussiansplattingrealtime}, more realistic and efficient rendering solutions have emerged. We integrate 3DGS rendering with established physical simulators such as MuJoCo~\cite{todorov2012mujoco} and Isaac Lab~\cite{mittal2023orbit}, enabling visually rich and physically accurate simulations. To this end, we develop \textit{\textbf{RoboSplatter}}, a 3DGS-based simulation rendering framework tailored for robotics simulation. As shown in Figure~\ref{fig:app2}, \textit{RoboSplatter} works seamlessly with MuJoCo to simulate robotic manipulation tasks, such as robotic arm grasping, while delivering high visual fidelity powered by 3DGS technology.

\section{Conclusion}
\label{sec:conclusion}

In this work, we present \textit{\textbf{{\mymethod}}}, the first comprehensive platform for interactive 3D world generation to the needs of embodied intelligence related research. Our system enables controllable and diverse creation of real-to-sim digital twins, alongside large-scale generation of 3D rigid and articulated object and 3D scene assets. These assets can be seamlessly integrated into various simulators such as OpenAI Gym\cite{openai_gym}, Isaac Lab\cite{mittal2023orbit}, MuJoCo\cite{todorov2012mujoco} and SAPIEN\cite{Xiang_2020_SAPIEN} for tasks such as ground-truth generation, evaluation, and reinforcement learning. The generated assets achieve state-of-the-art quality in both visual fidelity and physical realism, and are enriched with detailed annotations, including quality inspection labels, watertight geometry, and dual 3D representations in both 3DGS\cite{kerbl20233dgaussiansplattingrealtime} and mesh formats. To promote research and practical adoption, we release the pipeline as an open-source, user-friendly toolkit and service.

\clearpage

\begin{figure*}
\centering
\includegraphics[width=0.75\linewidth]
{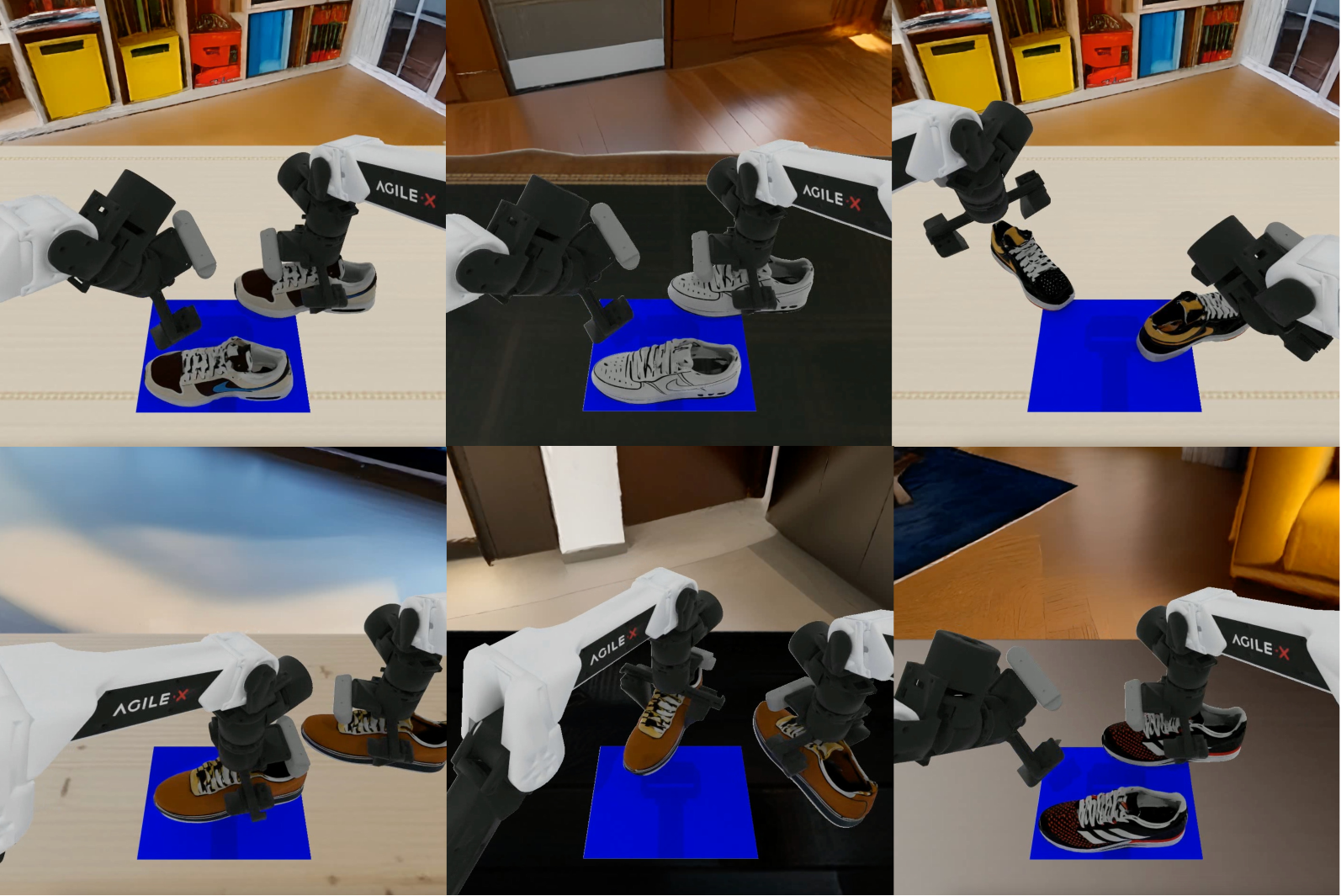}
\vspace*{-2mm}
\caption{
Interaction 3D World Generation with \textit{{\mymethod}}.
\textit{{\mymethod}} enables easy construction of diverse interactive 3D worlds for simulating and evaluating dual-arm shoe-grasping tasks in RoboTwin\cite{mu2025robotwin}.
}
\vspace*{-3mm}
\label{fig:shoes_scene}
\end{figure*}

\clearpage

%% file: main.bbl
\begin{thebibliography}{60}
\providecommand{\natexlab}[1]{#1}
\providecommand{\url}[1]{\texttt{#1}}
\expandafter\ifx\csname urlstyle\endcsname\relax
  \providecommand{\doi}[1]{doi: #1}\else
  \providecommand{\doi}{doi: \begingroup \urlstyle{rm}\Url}\fi

\bibitem[Achiam et~al.(2023)Achiam, Adler, Agarwal, Ahmad, Akkaya, Aleman, Almeida, Altenschmidt, Altman, Anadkat, et~al.]{achiam2023gpt}
Josh Achiam, Steven Adler, Sandhini Agarwal, Lama Ahmad, Ilge Akkaya, Florencia~Leoni Aleman, Diogo Almeida, Janko Altenschmidt, Sam Altman, Shyamal Anadkat, et~al.
\newblock Gpt-4 technical report.
\newblock \emph{arXiv preprint arXiv:2303.08774}, 2023.

\bibitem[AI(2023)]{briaai_rmbg_1_4}
BRIA AI.
\newblock Rmbg-1.4: Background removal model.
\newblock \url{https://huggingface.co/briaai/RMBG-1.4}, 2023.
\newblock Accessed: 2025-05-19.

\bibitem[Bensadoun et~al.(2024)Bensadoun, Monnier, Kleiman, Kokkinos, Siddiqui, Kariya, Harosh, Shapovalov, Graham, Garreau, Karnewar, Cao, Azuri, Makarov, Le, Toisoul, Novotny, Gafni, Neverova, and Vedaldi]{bensadoun2024meta3dgen}
Raphael Bensadoun, Tom Monnier, Yanir Kleiman, Filippos Kokkinos, Yawar Siddiqui, Mahendra Kariya, Omri Harosh, Roman Shapovalov, Benjamin Graham, Emilien Garreau, Animesh Karnewar, Ang Cao, Idan Azuri, Iurii Makarov, Eric-Tuan Le, Antoine Toisoul, David Novotny, Oran Gafni, Natalia Neverova, and Andrea Vedaldi.
\newblock Meta 3d gen, 2024.

\bibitem[Brockman et~al.(2016)Brockman, Cheung, Pettersson, Schneider, Schulman, Tang, and Zaremba]{openai_gym}
Greg Brockman, Vicki Cheung, Ludwig Pettersson, Jonas Schneider, John Schulman, Jie Tang, and Wojciech Zaremba.
\newblock Openai gym, 2016.

\bibitem[Charatan et~al.(2024)Charatan, Li, Tagliasacchi, and Sitzmann]{charatan2024pixelsplat}
David Charatan, Sizhe Li, Andrea Tagliasacchi, and Vincent Sitzmann.
\newblock pixelsplat: 3d gaussian splats from image pairs for scalable generalizable 3d reconstruction, 2024.

\bibitem[Chen et~al.(2024{\natexlab{a}})Chen, Xu, Zheng, Zhuang, Pollefeys, Geiger, Cham, and Cai]{Chen_2024_mvsplat}
Yuedong Chen, Haofei Xu, Chuanxia Zheng, Bohan Zhuang, Marc Pollefeys, Andreas Geiger, Tat-Jen Cham, and Jianfei Cai.
\newblock \emph{MVSplat: Efficient 3D Gaussian Splatting from Sparse Multi-view Images}, page 370–386.
\newblock Springer Nature Switzerland, 2024{\natexlab{a}}.

\bibitem[Chen et~al.(2024{\natexlab{b}})Chen, Walsman, Memmel, Mo, Fang, Vemuri, Wu, Fox, and Gupta]{chen2024urdformerpipelineconstructingarticulated}
Zoey Chen, Aaron Walsman, Marius Memmel, Kaichun Mo, Alex Fang, Karthikeya Vemuri, Alan Wu, Dieter Fox, and Abhishek Gupta.
\newblock Urdformer: A pipeline for constructing articulated simulation environments from real-world images, 2024{\natexlab{b}}.

\bibitem[Chung et~al.(2023)Chung, Lee, Nam, Lee, and Lee]{chung2023luciddreamerdomainfreegeneration3d}
Jaeyoung Chung, Suyoung Lee, Hyeongjin Nam, Jaerin Lee, and Kyoung~Mu Lee.
\newblock Luciddreamer: Domain-free generation of 3d gaussian splatting scenes, 2023.

\bibitem[Dai et~al.(2024)Dai, Wong, Jiang, Wang, Gokmen, Zhang, Wu, and Fei-Fei]{dai2024automatedcreationdigitalcousins}
Tianyuan Dai, Josiah Wong, Yunfan Jiang, Chen Wang, Cem Gokmen, Ruohan Zhang, Jiajun Wu, and Li Fei-Fei.
\newblock Automated creation of digital cousins for robust policy learning, 2024.

\bibitem[et~al.(2023)]{qwen}
Jinze~Bai et al.
\newblock Qwen technical report.
\newblock \emph{arXiv preprint arXiv:2309.16609}, 2023.

\bibitem[et~al.(2024)]{hong2024lrm}
Yicong~Hong et al.
\newblock Lrm: Large reconstruction model for single image to 3d, 2024.

\bibitem[et~al.(2025)]{zhao2025hunyuan3d20scalingdiffusion}
Zibo~Zhao et al.
\newblock Hunyuan3d 2.0: Scaling diffusion models for high resolution textured 3d assets generation, 2025.

\bibitem[Feng et~al.(2023)Feng, Liu, Cui, and Xie]{feng2023diffusion360seamless360degree}
Mengyang Feng, Jinlin Liu, Miaomiao Cui, and Xuansong Xie.
\newblock Diffusion360: Seamless 360 degree panoramic image generation based on diffusion models, 2023.

\bibitem[Gadre et~al.(2021)Gadre, Ehsani, and Song]{gadre2021act}
Samir~Yitzhak Gadre, Kiana Ehsani, and Shuran Song.
\newblock Act the part: Learning interaction strategies for articulated object part discovery.
\newblock In \emph{Proceedings of the IEEE/CVF International Conference on Computer Vision}, pages 15752--15761, 2021.

\bibitem[Gatis(2025)]{Gatis_rembg_2025}
Daniel Gatis.
\newblock rembg, 2025.
\newblock A tool to remove images background.

\bibitem[Ho et~al.(2020)Ho, Jain, and Abbeel]{ho2020denoising}
Jonathan Ho, Ajay Jain, and Pieter Abbeel.
\newblock Denoising diffusion probabilistic models.
\newblock \emph{Advances in neural information processing systems}, 33:\penalty0 6840--6851, 2020.

\bibitem[Huang et~al.(2024)Huang, Guo, Wang, Yi, Ma, Cao, and Sheng]{huang2024mvadapter}
Zehuan Huang, Yuan-Chen Guo, Haoran Wang, Ran Yi, Lizhuang Ma, Yan-Pei Cao, and Lu Sheng.
\newblock Mv-adapter: Multi-view consistent image generation made easy, 2024.

\bibitem[Katara et~al.(2023)Katara, Xian, and Fragkiadaki]{katara2023gen2sim}
Pushkal Katara, Zhou Xian, and Katerina Fragkiadaki.
\newblock Gen2sim: Scaling up robot learning in simulation with generative models, 2023.

\bibitem[Kerbl et~al.(2023)Kerbl, Kopanas, Leimkühler, and Drettakis]{kerbl20233dgaussiansplattingrealtime}
Bernhard Kerbl, Georgios Kopanas, Thomas Leimkühler, and George Drettakis.
\newblock 3d gaussian splatting for real-time radiance field rendering, 2023.

\bibitem[Kirillov et~al.(2023)Kirillov, Mintun, Ravi, Mao, Rolland, Gustafson, Xiao, Whitehead, Berg, Lo, Doll{\'a}r, and Girshick]{kirillov2023segany}
Alexander Kirillov, Eric Mintun, Nikhila Ravi, Hanzi Mao, Chloe Rolland, Laura Gustafson, Tete Xiao, Spencer Whitehead, Alexander~C. Berg, Wan-Yen Lo, Piotr Doll{\'a}r, and Ross Girshick.
\newblock Segment anything.
\newblock \emph{arXiv:2304.02643}, 2023.

\bibitem[Liu et~al.(2025)Liu, Iliash, Chang, Savva, and Mahdavi-Amiri]{liu2025singaposingleimagecontrolled}
Jiayi Liu, Denys Iliash, Angel~X. Chang, Manolis Savva, and Ali Mahdavi-Amiri.
\newblock Singapo: Single image controlled generation of articulated parts in objects, 2025.

\bibitem[Liu et~al.(2023{\natexlab{a}})Liu, Wu, Hoorick, Tokmakov, Zakharov, and Vondrick]{liu2023zero1to3}
Ruoshi Liu, Rundi Wu, Basile~Van Hoorick, Pavel Tokmakov, Sergey Zakharov, and Carl Vondrick.
\newblock Zero-1-to-3: Zero-shot one image to 3d object, 2023{\natexlab{a}}.

\bibitem[Liu et~al.(2023{\natexlab{b}})Liu, Wu, Hoorick, Tokmakov, Zakharov, and Vondrick]{liu2023zero1to3zeroshotimage3d}
Ruoshi Liu, Rundi Wu, Basile~Van Hoorick, Pavel Tokmakov, Sergey Zakharov, and Carl Vondrick.
\newblock Zero-1-to-3: Zero-shot one image to 3d object, 2023{\natexlab{b}}.

\bibitem[Liu et~al.(2023{\natexlab{c}})Liu, Xie, Liu, and Wong]{liu2023text}
Yuxin Liu, Minshan Xie, Hanyuan Liu, and Tien-Tsin Wong.
\newblock Text-guided texturing by synchronized multi-view diffusion.
\newblock \emph{arXiv preprint arXiv:2311.12891}, 2023{\natexlab{c}}.

\bibitem[{Meshy.ai}(2025)]{meshyai}
{Meshy.ai}.
\newblock Meshy.ai: Ai-powered 3d mesh generation, 2025.
\newblock Accessed: 2025-05-09.

\bibitem[Mittal et~al.(2023)Mittal, Yu, Yu, Liu, Rudin, Hoeller, Yuan, Singh, Guo, Mazhar, Mandlekar, Babich, State, Hutter, and Garg]{mittal2023orbit}
Mayank Mittal, Calvin Yu, Qinxi Yu, Jingzhou Liu, Nikita Rudin, David Hoeller, Jia~Lin Yuan, Ritvik Singh, Yunrong Guo, Hammad Mazhar, Ajay Mandlekar, Buck Babich, Gavriel State, Marco Hutter, and Animesh Garg.
\newblock Orbit: A unified simulation framework for interactive robot learning environments.
\newblock \emph{IEEE Robotics and Automation Letters}, 8\penalty0 (6):\penalty0 3740--3747, 2023.

\bibitem[Mo et~al.(2021)Mo, Guibas, Mukadam, Gupta, and Tulsiani]{mo2021where2act}
Kaichun Mo, Leonidas~J Guibas, Mustafa Mukadam, Abhinav Gupta, and Shubham Tulsiani.
\newblock Where2act: From pixels to actions for articulated 3d objects.
\newblock In \emph{Proceedings of the IEEE/CVF International Conference on Computer Vision}, pages 6813--6823, 2021.

\bibitem[Mu et~al.(2025)Mu, Chen, Peng, Chen, Gao, Zou, Lin, Xie, and Luo]{mu2025robotwin}
Yao Mu, Tianxing Chen, Shijia Peng, Zanxin Chen, Zeyu Gao, Yude Zou, Lunkai Lin, Zhiqiang Xie, and Ping Luo.
\newblock Robotwin: Dual-arm robot benchmark with generative digital twins (early version), 2025.

\bibitem[Nichol and Dhariwal(2021)]{nichol2021improved}
Alexander~Quinn Nichol and Prafulla Dhariwal.
\newblock Improved denoising diffusion probabilistic models.
\newblock In \emph{International conference on machine learning}, pages 8162--8171. PMLR, 2021.

\bibitem[Poole et~al.(2022{\natexlab{a}})Poole, Jain, Barron, and Mildenhall]{poole2022dreamfusion}
Ben Poole, Ajay Jain, Jonathan~T. Barron, and Ben Mildenhall.
\newblock Dreamfusion: Text-to-3d using 2d diffusion, 2022{\natexlab{a}}.

\bibitem[Poole et~al.(2022{\natexlab{b}})Poole, Jain, Barron, and Mildenhall]{poole2022dreamfusiontextto3dusing2d}
Ben Poole, Ajay Jain, Jonathan~T. Barron, and Ben Mildenhall.
\newblock Dreamfusion: Text-to-3d using 2d diffusion, 2022{\natexlab{b}}.

\bibitem[Pu et~al.(2024)Pu, Zhao, and Lian]{Pu_2024}
Guo Pu, Yiming Zhao, and Zhouhui Lian.
\newblock Pano2room: Novel view synthesis from a single indoor panorama.
\newblock In \emph{SIGGRAPH Asia 2024 Conference Papers}, page 1–11. ACM, 2024.

\bibitem[Qian and Fouhey(2023)]{qian2023understanding}
Shengyi Qian and David~F Fouhey.
\newblock Understanding 3d object interaction from a single image.
\newblock In \emph{Proceedings of the IEEE/CVF International Conference on Computer Vision}, pages 21753--21763, 2023.

\bibitem[Radford et~al.(2019)Radford, Wu, Child, Luan, Amodei, and Sutskever]{radford2019language}
Alec Radford, Jeff Wu, Rewon Child, David Luan, Dario Amodei, and Ilya Sutskever.
\newblock Language models are unsupervised multitask learners.
\newblock 2019.

\bibitem[Radford et~al.(2021)Radford, Kim, Hallacy, Ramesh, Goh, Agarwal, Sastry, Askell, Mishkin, Clark, Krueger, and Sutskever]{clip2021}
Alec Radford, Jong~Wook Kim, Chris Hallacy, Aditya Ramesh, Gabriel Goh, Sandhini Agarwal, Girish Sastry, Amanda Askell, Pamela Mishkin, Jack Clark, Gretchen Krueger, and Ilya Sutskever.
\newblock Learning transferable visual models from natural language supervision, 2021.

\bibitem[Richardson et~al.(2023)Richardson, Metzer, Alaluf, Giryes, and Cohen-Or]{texture}
Elad Richardson, Gal Metzer, Yuval Alaluf, Raja Giryes, and Daniel Cohen-Or.
\newblock Texture: Text-guided texturing of 3d shapes, 2023.

\bibitem[Rombach et~al.(2022)Rombach, Blattmann, Lorenz, Esser, and Ommer]{ldm}
Robin Rombach, Andreas Blattmann, Dominik Lorenz, Patrick Esser, and Bj{\"o}rn Ommer.
\newblock High-resolution image synthesis with latent diffusion models.
\newblock In \emph{Proceedings of the IEEE/CVF conference on computer vision and pattern recognition}, pages 10684--10695, 2022.

\bibitem[Schuhmann(2025)]{schuhmann_aesthetic_2025}
Christoph Schuhmann.
\newblock Aesthetic subsets in laion 2170337258 samples, 2025.
\newblock Retrieved May 16, 2025.

\bibitem[Shi et~al.(2024)Shi, Wang, Ye, Long, Li, and Yang]{shi2024mvdream}
Yichun Shi, Peng Wang, Jianglong Ye, Mai Long, Kejie Li, and Xiao Yang.
\newblock Mvdream: Multi-view diffusion for 3d generation, 2024.

\bibitem[Team(2024)]{kolors}
Kolors Team.
\newblock Kolors: Effective training of diffusion model for photorealistic text-to-image synthesis.
\newblock \emph{arXiv preprint}, 2024.

\bibitem[Team(2025)]{hunyuan3d22025tencent}
Tencent~Hunyuan3D Team.
\newblock Hunyuan3d 2.0: Scaling diffusion models for high resolution textured 3d assets generation, 2025.
\newblock \url{https://huggingface.co/tencent/Hunyuan3D-2/tree/main/hunyuan3d-delight-v2-0}.

\bibitem[Todorov et~al.(2012)Todorov, Erez, and Tassa]{todorov2012mujoco}
Emanuel Todorov, Tom Erez, and Yuval Tassa.
\newblock Mujoco: A physics engine for model-based control.
\newblock In \emph{2012 IEEE/RSJ International Conference on Intelligent Robots and Systems}, pages 5026--5033. IEEE, 2012.

\bibitem[Wang et~al.(2023)Wang, Wang, Chen, Wang, Loy, and Liu]{wang2023perfpanoramicneuralradiance}
Guangcong Wang, Peng Wang, Zhaoxi Chen, Wenping Wang, Chen~Change Loy, and Ziwei Liu.
\newblock Perf: Panoramic neural radiance field from a single panorama, 2023.

\bibitem[Wang et~al.()Wang, Xie, Dong, and Shan]{wang2021realesrgan}
Xintao Wang, Liangbin Xie, Chao Dong, and Ying Shan.
\newblock Real-esrgan: Training real-world blind super-resolution with pure synthetic data.
\newblock In \emph{International Conference on Computer Vision Workshops (ICCVW)}.

\bibitem[Wang et~al.(2025)Wang, Tang, Gan, Fox, Mo, Narang, and Akinola]{wang2025matchmaker}
Yian Wang, Bingjie Tang, Chuang Gan, Dieter Fox, Kaichun Mo, Yashraj Narang, and Iretiayo Akinola.
\newblock Matchmaker: Automated asset generation for robotic assembly, 2025.

\bibitem[Wu et~al.(2024{\natexlab{a}})Wu, Liu, Cai, Yan, Wang, Hu, Duan, and Ma]{wu2024unique3d}
Kailu Wu, Fangfu Liu, Zhihan Cai, Runjie Yan, Hanyang Wang, Yating Hu, Yueqi Duan, and Kaisheng Ma.
\newblock Unique3d: High-quality and efficient 3d mesh generation from a single image, 2024{\natexlab{a}}.

\bibitem[Wu et~al.(2024{\natexlab{b}})Wu, Chen, Yang, Guo, Li, and Zhang]{wu2024lamp}
Ruiqi Wu, Liangyu Chen, Tong Yang, Chunle Guo, Chongyi Li, and Xiangyu Zhang.
\newblock Lamp: Learn a motion pattern for few-shot video generation.
\newblock In \emph{Proceedings of the IEEE/CVF Conference on Computer Vision and Pattern Recognition}, pages 7089--7098, 2024{\natexlab{b}}.

\bibitem[Wu et~al.(2025)Wu, Wang, Liu, Guo, Qiu, Li, Huang, Su, and Cheng]{wu2025dipodualstateimagescontrolled}
Ruiqi Wu, Xinjie Wang, Liu Liu, Chunle Guo, Jiaxiong Qiu, Chongyi Li, Lichao Huang, Zhizhong Su, and Ming-Ming Cheng.
\newblock Dipo: Dual-state images controlled articulated object generation powered by diverse data, 2025.

\bibitem[Xiang et~al.(2020)Xiang, Qin, Mo, Xia, Zhu, Liu, Liu, Jiang, Yuan, Wang, Yi, Chang, Guibas, and Su]{Xiang_2020_SAPIEN}
Fanbo Xiang, Yuzhe Qin, Kaichun Mo, Yikuan Xia, Hao Zhu, Fangchen Liu, Minghua Liu, Hanxiao Jiang, Yifu Yuan, He Wang, Li Yi, Angel~X. Chang, Leonidas~J. Guibas, and Hao Su.
\newblock {SAPIEN}: A simulated part-based interactive environment.
\newblock In \emph{The IEEE Conference on Computer Vision and Pattern Recognition (CVPR)}, 2020.

\bibitem[Xiang et~al.(2024)Xiang, Lv, Xu, Deng, Wang, Zhang, Chen, Tong, and Yang]{xiang2024structured}
Jianfeng Xiang, Zelong Lv, Sicheng Xu, Yu Deng, Ruicheng Wang, Bowen Zhang, Dong Chen, Xin Tong, and Jiaolong Yang.
\newblock Structured 3d latents for scalable and versatile 3d generation.
\newblock \emph{arXiv preprint arXiv:2412.01506}, 2024.

\bibitem[Xie(2025)]{worldgen2025ziyangxie}
Ziyang Xie.
\newblock Worldgen: Generate any 3d scene in seconds.
\newblock \url{https://github.com/ZiYang-xie/WorldGen}, 2025.

\bibitem[Xu et~al.(2024)Xu, Shi, Yifan, Chen, Yang, Peng, Shen, and Wetzstein]{xu2024grm}
Yinghao Xu, Zifan Shi, Wang Yifan, Hansheng Chen, Ceyuan Yang, Sida Peng, Yujun Shen, and Gordon Wetzstein.
\newblock Grm: Large gaussian reconstruction model for efficient 3d reconstruction and generation, 2024.

\bibitem[Yang et~al.(2025)Yang, Tan, Zhang, Wu, Li, Wetzstein, Liu, and Lin]{yang2025layerpano3dlayered3dpanorama}
Shuai Yang, Jing Tan, Mengchen Zhang, Tong Wu, Yixuan Li, Gordon Wetzstein, Ziwei Liu, and Dahua Lin.
\newblock Layerpano3d: Layered 3d panorama for hyper-immersive scene generation, 2025.

\bibitem[Zeng et~al.(2024)Zeng, Chen, Qi, Liu, Zhao, Wang, Fu, Liu, and Yu]{zeng2024paint3d}
Xianfang Zeng, Xin Chen, Zhongqi Qi, Wen Liu, Zibo Zhao, Zhibin Wang, Bin Fu, Yong Liu, and Gang Yu.
\newblock Paint3d: Paint anything 3d with lighting-less texture diffusion models.
\newblock In \emph{Proceedings of the IEEE/CVF Conference on Computer Vision and Pattern Recognition}, pages 4252--4262, 2024.

\bibitem[Zhang et~al.(2023)Zhang, Rao, and Agrawala]{zhang2023adding}
Lvmin Zhang, Anyi Rao, and Maneesh Agrawala.
\newblock Adding conditional control to text-to-image diffusion models, 2023.

\bibitem[Zhang et~al.(2024{\natexlab{a}})Zhang, Wang, Zhang, Qiu, Pang, Jiang, Yang, Xu, and Yu]{zhang2024clay}
Longwen Zhang, Ziyu Wang, Qixuan Zhang, Qiwei Qiu, Anqi Pang, Haoran Jiang, Wei Yang, Lan Xu, and Jingyi Yu.
\newblock Clay: A controllable large-scale generative model for creating high-quality 3d assets, 2024{\natexlab{a}}.

\bibitem[Zhang et~al.(2024{\natexlab{b}})Zhang, Wang, Zhang, Qiu, Pang, Jiang, Yang, Xu, and Yu]{zhang2024claycontrollablelargescalegenerative}
Longwen Zhang, Ziyu Wang, Qixuan Zhang, Qiwei Qiu, Anqi Pang, Haoran Jiang, Wei Yang, Lan Xu, and Jingyi Yu.
\newblock Clay: A controllable large-scale generative model for creating high-quality 3d assets, 2024{\natexlab{b}}.

\bibitem[Zhang et~al.(2024{\natexlab{c}})Zhang, Liu, Xie, Yang, Liu, Yang, Zhang, Kou, Lin, Wang, and Jin]{zhang2024dreammat}
Yuqing Zhang, Yuan Liu, Zhiyu Xie, Lei Yang, Zhongyuan Liu, Mengzhou Yang, Runze Zhang, Qilong Kou, Cheng Lin, Wenping Wang, and Xiaogang Jin.
\newblock Dreammat: High-quality pbr material generation with geometry- and light-aware diffusion models, 2024{\natexlab{c}}.

\bibitem[Zhou et~al.(2024{\natexlab{a}})Zhou, Cheng, Yu, Tian, and Yuan]{zhou2024holodreamerholistic3dpanoramic}
Haiyang Zhou, Xinhua Cheng, Wangbo Yu, Yonghong Tian, and Li Yuan.
\newblock Holodreamer: Holistic 3d panoramic world generation from text descriptions, 2024{\natexlab{a}}.

\bibitem[Zhou et~al.(2024{\natexlab{b}})Zhou, Fan, Xu, Chang, Chari, Bharadwaj, You, Wang, and Kadambi]{zhou2024dreamscene360unconstrainedtextto3dscene}
Shijie Zhou, Zhiwen Fan, Dejia Xu, Haoran Chang, Pradyumna Chari, Tejas Bharadwaj, Suya You, Zhangyang Wang, and Achuta Kadambi.
\newblock Dreamscene360: Unconstrained text-to-3d scene generation with panoramic gaussian splatting, 2024{\natexlab{b}}.

\end{thebibliography}
